\documentclass[10pt,journal,compsoc]{IEEEtran}
 
\ifCLASSOPTIONcompsoc
  \usepackage[nocompress]{cite}
\else 
  \usepackage{cite}
\fi

  \usepackage[pdftex]{graphicx}
  \graphicspath{{./images/}{./charts/}{./pdfgraphics/}}
  \DeclareGraphicsExtensions{.png,.pdf,.jpeg}

\ifCLASSOPTIONcompsoc
  \usepackage[caption=false,font=footnotesize,labelfont=sf,textfont=sf]{subfig}
\else
  \usepackage[caption=false,font=footnotesize]{subfig}
\fi

\usepackage{dblfloatfix}
\usepackage{url}

\usepackage[utf8]{inputenc}
\usepackage{bm}
\usepackage{makeidx}  % allows for indexgeneration
\usepackage{color}
\usepackage{tabularx}
\usepackage{multirow}
\usepackage{rotating}

\begin{document}

\title{Evaluating and Characterizing Incremental Learning from Non-Stationary Data}

\author{Alejandro~Cervantes,
        Christian~Gagn\'{e},
        Pedro~Isasi,
        and~Marc~Parizeau% <-this % stops a % space
\IEEEcompsocitemizethanks{\IEEEcompsocthanksitem Alejandro~Cervantes and
Pedro~Isasi are with the Department of Computer Science, University Carlos III
de Madrid, Legan\'{e}s, Madrid 29911, Spain, e-mail:
\{alejandro.cervantes,pedro.isasi\}@uc3m.es
\IEEEcompsocthanksitem Christian Gagn\'{e} and Marc Parizeau are with
Laboratoire de Vision et Syst\`{e}mes Num\'{e}riques (LVSN), Universit\'{e} Laval, Qu\'{e}bec
 (Qu\'{e}bec), Canada~~G1V 0A6,
 e-mail:\{christian.gagne,marc.parizeau\}@gel.ulaval.ca}% <-this % stops a space
}

\IEEEtitleabstractindextext{%

\begin{abstract}
Incremental learning from non-stationary data poses special
challenges to the field of machine learning. Although new algorithms
have been developed for this, assessment of results and
comparison of behaviors are still open problems, mainly because 
evaluation metrics, adapted from more traditional tasks,
can be ineffective in this context. 
Overall, there is a lack of common testing practices. This paper thus
presents a testbed for incremental non-stationary learning algorithms, 
based on specially designed synthetic datasets. Also, test results are reported for 
some well-known algorithms to show that the proposed methodology is effective at 
characterizing their strengths and weaknesses. It is expected that this methodology 
will provide a common basis for evaluating future contributions in the field.
\end{abstract}

}
\maketitle

\IEEEdisplaynontitleabstractindextext

\IEEEpeerreviewmaketitle

\IEEEraisesectionheading{\section{Introduction}\label{sec:introduction}}

\IEEEPARstart{N}{etworked} computing devices are now ubiquitous. The processing
capacity and size of these devices allow them to be used outside the office, allowing them to deal almost in real-time with the enormous amount of data that can be sensed from physical or virtual environments. These devices will be used more and more to process large scale continuous flows of data using machine learning techniques. In such a context, we clearly need methods for achieving Incremental Learning (IL), that is the capacity to process the data by learning methods in an incremental fashion, dealing with the changes that may occur  in the environment we are sensing.

The notion of IL has been previously used in different contexts.
Giraud-Carrier~\cite{Giraud00} distinguishes the concept of an Incremental Learning
Task (ILT) from the concept of an Incremental Learning Algorithm (ILA). An ILT
is defined as: ``\textit{A learning task where the training examples used to
solve it become available over time, usually one at a time.}'', whereas an ILA
``\textit{is incremental if, for any given training sample, it produces a
sequence of hypotheses such that the current hypothesis depends only on the
previous hypotheses and the current example.}''. This definition implies that
the task is essentially sequential, and that the algorithm implements some type
of Markov process.

Some application domains that may be considered ILTs, or at least, where ILAs are usually considered well suited are:

\begin{itemize}
       
     \item \textbf{Learning with Concept Drift}~\cite{Helmbold94}.
       Concept drift appears when new data is not consistent with past data. 
       This means that some unobserved aspect of reality was correlated with
       expected results, and varying values of that ``hidden context'' must
       force variation on learned models. Most references in this field assume
       a principle of batch learning, that is, complete training data is available and can be
       processed before the algorithm is requested to provide an answer.
       
    \item \textbf{Learning from Large Datasets}~\cite{bottou08b}. Defined as scenarios
       where a huge amount of learning data is available, which imposes constraints on
       the space required for representation. 

    \item \textbf{Learning from Data Streams}. Defined by a continuous 
       stream of learning data, which makes preprocessing of data
       unfeasible. These problems are also assumed to include the restriction
       that data can only be read once (one-pass learning) and usually assume
       non-stationarity of data
       %~\cite{Aggarwal2003}. 
       A recent review of current
       techniques in this field can be found in~\cite{Khol10}.

    \item \textbf{Real-time Learning}. A particular case of Learning from Data Streams
       where hard time constraints are also imposed. The learner has to be
       ready to respond in a specific (usually reduced) period of time. This
       may also require special-purpose hardware for many applications. See~\cite{DomHul00}
       for some specifics in this field.

\end{itemize}

An early definition from Aha and Kibler~\cite{Aha91} limits IL to sequential processing of
examples in an instance based learning algorithm. This restrictive definition,
however, provides little insight on the key issues of IL. Other more detailed
definitions~\cite{ElwellP11,StreetSEA01,Gama09,MartinezRego2011} include a
mixture of task features that incremental algorithms need to address. They vary
slightly depending on the topic of the paper, for instance in~\cite{StreetSEA01}
it is related to Learning from Large Datasets, while in~\cite{Gama09} they are
referring to Learning from Data Streams.

For the purposes of the present work, we are using the following three task
features to define an ILT: Sequential Data (SD), Continuous Flow (CF), and
Non-Stationary Distributions (NSD). SD simply means that data becomes available
in a certain meaningful order. CF states that a never ending flow of new
training data need to be learned. NSD specifies that underlying class
distributions and priors may evolve over time, and thus that embedded class
models should be able to adapt. An algorithm that deals with SD and CF, but not
NSD, is often called \emph{online}, whereas an algorithm that assumes that all
training data is available before learning is called \emph{offline}. An ILA is
simply an algorithm capable of dealing with all three features of an ILT.

For the NSD feature, any change in class distributions can be assimilated to a
concept drift. Typical class instances may thus move within the feature space,
and their spread (i.e., variance of their distribution) can also evolve dynamically. But the ILT
is more general than just a concept drift. Changes in class priors allow for dynamical appearance or
disappearance of concepts.

Similarly, an ILA is also characterized by the following four algorithm
features: Sequential Processing (SP), Online Processing (OP), Any-Time Response
(ATR), and Adaptation to Change (AC). SP means that data is learned as it arrives,
thus taking data ordering into account. OP implies that a complete or final data
set is never expected; data should be processed in a single pass. ATR states
that a prediction of the data class label must be available at any time.
Finally, AC specifies that embedded class models must be able to adapt to
changes in data, either stationary or non-stationary.

Other constraints arise from CF and OP: Time Invariance (TI), and Space
Invariance (SI). TI means that each training sample should be processed in
constant time, regardless of the amount of data encountered so-far.
SI implies that internal data structures should require an approximately constant amount of
memory, also independent of the amount of previously processed data.

A typical ILA will process patterns either one at a time or in chunks, creating
a new model from the past model or models and new information at some intervals.
In the meantime, it is able to respond with the ``best guess'' answer of the
moment from the ``current'' model.

Table~\ref{tab:areasVsILT} summarizes how different
fields of application can be classified with respect to task and algorithm features, and
specifies the particular \emph{incremental learning from non-stationary data} task, 
which is the scope of the rest of the paper.

\begin{table*}
\centering
\caption{Relation between areas of research in incremental learning and the various tasks, algorithms features, and constraints they consider (SD: Sequential Data, CF: Continuous Flow, NSD: Non-Stationary Distributions, SP: Sequential Processing, OP: Online Processing, ATR: Any-Time Response, AC: Adaptation to Change, TI: Time Invariance, and SI: Space Invariance).}
\label{tab:areasVsILT}
%\resizebox{\linewidth}{!}{
\begin{tabular}{l|ccc|cccc|cc}
\hline
Area                                 & SD     & CF     & NSD     & SP   & OP     & ATR    & AC      & TI     & SI \\ 
\hline
Learning with concept drift          &  Yes   & No     & Yes     & No   & No     & No     & Yes     & Weak   & Weak   \\ 
Learning from large datasets         &  No    & Yes    & No      & No   & Yes   
& Yes    & No      & Weak     & Hard     \\
Learning from data streams           &  Yes   & Yes    & Usually & Yes  & Yes    & Yes    & Usually & Weak   & Hard   \\ 
Real-time learning from data streams &  Yes   & Yes    & Usually & Yes  & Yes    & Yes    & Usually & Hard   & Hard   \\ 
\hline
Learning from non-stationary data    &  Yes   & Yes    & Yes     & Yes  & Yes    & Yes    & Yes     & Weak   & Weak   \\ 
\hline
\end{tabular}
%}
%\medskip
\end{table*}

\section{IL with Non-Stationary Environments}

The IL domain is vast and, as illustrated in the previous section, its scope varies according to the author. In the current work, we are restricting our review to the specific literature related to incremental learning from non-stationary data, as represented in Table~\ref{tab:areasVsILT}. More specifically, we are reviewing literature of related fields, analyzing the type of algorithms that are being used, the datasets used to test them, and the measurements used to evaluate and compare algorithms on those datasets. 

\subsection{Algorithms for Incremental Non-Stationary Learning}

One of the first models that directly addresses many of the concerns specific
to ILA was Adaptive Resonance Theory (ART~\cite{CarGroART88}). The authors 
address a core issue for non-stationary systems in what they
call the stability versus plasticity dilemma: learners must be
able to react to significant events (plasticity) but not every event
has to be learned (stability), as this can be proven to lead to
unstable behavior. ART deals with the problem by being able
to change between two different ``learning'' modes, and
autonomously recognize when the algorithm should change between them.

This key issue is the concern of most of the algorithms that
are specifically designed to deal with non-stationary data in an
incremental manner. However, the underlying representation may vary, from
classification trees to neural network or support
vector approaches.

One of the first attempts at IL is IB3~\cite{Aha91}.
In this case representation is a set of instances that are considered
``relevant'' and used with a nearest neighbor classification rule. 
Incremental learning is based on an adaptive learning and forgetting mechanism: 
it keeps or deletes instances based on the accuracy of the instance compared to the
relative frequency of its class. As we will see later, determining
which instances are relevant is a common way of ``incrementalization'' of
algorithms whose underlying models are more complex. Nearest
neighbor approaches have been studied in~\cite{AlippiR08}, where
the authors suggest a solution based on an instance weighting approach.
If a proper window size can be selected to only consider the most recent
patterns, very simple lazy learning approaches can also be competitive~\cite{BifetPha13}.

In tree learning, an algorithm that is often referenced is CVFDT~\cite{Hulten01CVFDT}.
It is a decision tree classifier that is able to rapidly process new data incrementally, thus
making it adequate for learning from real time data streams. It is also able
to learn from non-stationary data, using a sliding window of stored
patterns. The mechanism updates the current tree when new data arrives, learning the new pattern and
forgetting the oldest one. When this operation invalidates parts of the
current model, one or more candidate replacements are generated; when
the accuracy of one of the replacements exceeds the accuracy of the current
model, that part of the model is replaced by the candidate. The window size changes
dynamically during the execution.

Neural network learning can usually be performed in an incremental way. For
instance, in~\cite{MartinezRego2011} an incremental online learning method
for single layer Neural Networks that is able to balance learning from past and
new instances, is presented. The method is based on an incremental weight adapting that
was previously proven to be lossless, that is, equivalent
to batch learning up to the same data. The advantage is that this network does
not include an explicit change detection mechanism and does not store past models.
The core of the model is a forgetting function incorporated into the objective function
to be minimized by the weight adapting mechanism. Selection of the forgetting function
becomes the key to efficient performance in non-stationary environments.

A field that has increasing interest in non-stationary
environments is ensemble based learning. Several reviews of the specific
mechanisms used in this field are available (see for instance~\cite{MinYao12});
one of the
basic reasons to use ensembles is that the incremental and adaptive part of the algorithm
can be achieved independently of the nature of the base classifier used.

One of the first attempts on this subject can be found in~\cite{StreetSEA01}.
In this case, classifiers are built based on small chunks of training data; when the ensemble is
complete, new classifiers can replace old ones based on the accuracy increase
over the new chunks of data. This allows the complete ensemble to self-adapt
to changing data.

A typical mechanism for ensemble learning is dynamic combination. For instance,
this is used in Learn++~\cite{MuhlPol07} to learn incrementally in the context of
non-stationary data. The ensemble dynamically combines classifiers based on weights 
that depend on the success rate of each classifier for the current dataset.
So this weight allocation is the part of the algorithm that is incremental and
able to adapt to non-stationary data. In recent versions of this ensemble
approach, such as Learn++.NSE~\cite{ElwellP11}, ensembles that had
% FOR TKDE Removed ,~\cite{KarMuhl08}
good accuracy in the past are not deleted, so they can become active in the
future. In this way, if the nature of the change is cyclical, the algorithm will
have an advantage when past conditions reappear. Another aspect of incremental
learning is considered in~\cite{Muhl09learnNC}, that has a better behavior in
situations where new classes appear in data.

Also new data can be used to update the underlying classifiers, using mechanisms
derived from the bagging or boosting batch learning algorithms.
Online bagging or boosting were proposed by Oza~\cite{oza01}; and
variants have been proposed that take into account non-stationary data, such as
ONSBoost~\cite{pocock1ONSBoost10}.
An analysis of several versions of online bagging or boosting is performed in ~\cite{bifet09}.

\subsection{Incrementalization of Algorithms}

Renewed interest on incremental algorithms has led to 
a significant trend in research that tries to adapt existing non-incremental algorithms to
an incremental behavior. 

This incrementalization approach can be conducted in several ways:

\begin{itemize}
\item Adapt the learning mechanism itself so it becomes purely incremental. For instance,
weights on a Neural Network are adapted every time a new pattern is available to the system.
\item Use an instance windows or instance weighting mechanism but make no modifications
to the algorithm itself; recalculate model at fixed intervals
using these windows or weights to decide which data is used for training the new model.
\item Use all or part of the data to create an initial model; monitor changes in data 
(using a detection function) and rebuild the model whenever deemed necessary based
on new data.
\end{itemize}

Several studies have adapted non-incremental classification methods in one
of the above ways:

\begin{itemize}

\item Basic lazy learning algorithms, such as K-Nearest Neighbor can be
transformed into robust non-stationary classifiers provided they perform
classification using only a certain number of the most recent patterns received
for training (window size, $wsize$). If this parameter is well suited for a given
problem, this approach can outperform some popular NSL methods~\cite{BifetPha13}.
% FOR TKDE Remove ReadBif12

\item Evolutionary algorithms may adapt to change if a change detection mechanism
is implemented and used to vary their parameters. For instance in~\cite{RiekAGP09}
an Adaptive Genetic Programming classifier is used for non-stationary data. In this
case, data is divided in separate windows, either overlapping or non-overlapping. For
every window change, a number of iterations is required in order to allow the 
algorithm to converge to a solution.

\item Other classifiers based on an incremental concept use an incremental version of
Growing Neural Gas (GNG) as a way of positioning prototypes. GNG is a clustering
method that aims to learn a topology. Its behavior is unsupervised, but in 
Supervised Neural Gas~\cite{Fritzke94fastlearning}, GNG units
may be used as centers for a Radial Basis Function Neural Network to be applied
to classification. In~\cite{Prudent05} it was shown how GNG can adapt to slow changes due to its
forgetting mechanism, so this feature may be used to implement
adaptation to change. %$AF_4$

\item Incremental algorithms have been proposed for vector quantization methods,
including Incremental Learning Vector Quantization~\cite{XuOlvq09},
that is claimed to work in non-stationary environments. % ($AF_4$).

\item Online Support Vector Machine (OSVM) is usually oriented to provide
efficient SVM implementations able to deal with large datasets. Depending
on implementation, algorithms are able to perform sequential and online processing, and
deliver anytime response, but the SVM model is not able to adapt to change. 
%$AM_1$ to $AM_3$ but not $AM_4$.
In~\cite{CauExactOSVM00} an exact version of support vector learning is documented, 
that provides optimal adjustment of the SVM support vectors given a new sample.
In~\cite{Laskov06incrementalsupport} some implementation issues are studied in
order to provide an efficient support vector learning algorithm.

%In~\cite{zhengSVM10} learning is performed incrementally so the model may deal with large datasets.

\item Stochastic Gradient Descent (SGD) techniques have been proven in~\cite{bottou10} 
to be successful options for online classification of large datasets using SVMs. The accuracy
loss due to its approximate nature (compared to exact optimization techniques) is easily balanced
by its computational effectiveness. The method can be incremental as SGD does not
remember which examples were used to adjust the classifier in previous iterations. SGD update
functions are provided for several classification and regression algorithms, including 
Adaline, Perceptron, and SVMs.

\end{itemize}

\subsection{Algorithms Dealing with Concept Drift}

The field of Concept Drift (CD) explicitly deals with non-stationary data,
while the incremental nature of learning is not required. Most work is related to algorithms
that work on complete training datasets in batch mode. However, there is a trend
that is moving towards considering how mechanisms might work on online scenarios. 
A comprehensive review of the field can be found in~\cite{Tsymbal04}.

A learning task with CD can be characterized in terms of two
orthogonal dimensions: drift frequency~\cite{Kuh91} and drift
extent~\cite{Helmbold91}. The authors work with bounding those dimensions
in order to ensure a certain degree of learning ability from a non-stationary
flow of data.

Widmer and Kubat \cite{Widmer96FLORA} introduce several important topics in
this area and propose some techniques to deal with them. This article provides a set of algorithms, called
the FLORA family, each version improving a specific aspect of the problem.
The basic mechanisms these algorithms use are an adaptive window
of trusted data and a repository of hypotheses. The algorithm monitors the
appearance of concept drifts by tracking changes in accuracy.

A family of algorithms based on Info-Fuzzy networks is formed by OLIN, IOLIN and Enhanced and
Multiple Model IOLIN~\cite{Last02,cohen2005eiolin,Cohen2008rtiolin}.
The first algorithm uses a regenerative
approach, constructing new models for each block of data (defined by a sliding window). The second
adopts an incremental approach whenever no major concept drift is detected between windows. The third
work introduces further enhancements and a way of recovering old models that become reusable.

In~\cite{Klink04}, the authors carry out an extensive treatment of concept drift using
non-incremental classifiers (SVM), but adopting a weighting schema for patterns
used for learning. Several cases are examined, ranging from sudden change to
slow drift. The authors claim that incremental versions of SVM were not of
interest as they are not designed for non-stationary data.

% See items 25 to 30 in bibliography of~\cite{Klink04}
%on all data? The problem is that support vectors are a sufficient description of the decision boundary
%between examples (P(y|x)), but not of the examples themselves (P(x)). Since there are typically only
%very few support vectors, their influence on the decision function in the next incremental learning step
%may be very small, if the data is distributed differently. The robustness of SVMs against outliers, usually
%a desired property, may now lead to not taking the old support vectors sufficiently into account for the
%new decision function

%One of the common problems with this type of straightforward adaptations is that
%they reduce the issue to a time-dependant global optimization of the optimal instance
%set for training. This is not the complete problem from our point of view.

%An important issue is to deal with local, not global, change. It is certain
%that not all knowledge will be subject to change in a given problem, and 
%that suggests that global mechanism will forget useful knowledge because of
%changes in recent data that is unrelated to the former.

\subsection{Evaluation Metrics}

Evaluating the performance of ILA presents
specific problems and must take into account aspects not present in
conventional evaluation of learning.

Early papers such as~\cite{Aha91} already described measures of performance
that are aimed to address each of the significant characteristics of so-called ``incremental 
algorithms''. In that work, measures that should be taken into account when
evaluating the success of a ILA are:

\begin{enumerate}
\item \textbf{Generality}: This is the kind of concept which can be described by the representation mechanism and is learnable by the algorithm.
\item \textbf{Accuracy}: This is the classification accuracy, either in terms of
averaged success or error rate.
\item \textbf{Learning Rate}: This is the speed at which classification accuracy increases during training. 
It is a more useful indicator of the performance of the learning algorithm than the accuracy for finite-sized training sets.
\item \textbf{Incorporation Costs}: These are the costs incurred while updating the concept descriptions with
a single training instance. They include classification costs.
\item \textbf{Storage Requirement}: This is the size of the concept description (for instance for instance-based algorithms,
it is defined as the number of saved instances used for classification decisions).
\end{enumerate}

Measures of accuracy and learning rate can be extended
for non-stationary data scenarios. However, accuracy has to be measured in a
specific way due to the continuous nature of learning, because
traditional accuracy measures
assume that: a) the dataset is finite; b) the concepts represented by learning do not change. 
We shall use learning rate to describe how the algorithm reacts to changes in data.

The continuous nature of input data 
means that common ways of estimating accuracy of learning cannot be used.
This applies to train-and-test, cross-validation, or leave-one-out methods, 
which are common practice in batch learning, where a representative test set can
be separated from training data at the beginning of the learning phase
and used only for validation of the generated model after training.

Almost every published paper that deals with non-stationary data evaluates
its performance by graphically comparing accuracy plots, either using an
accumulated accuracy or displaying average accuracy over a sliding window
of fixed size, to be able to account for transient states where algorithms
are adapting to changes.

Two different methods have been used to evaluate accuracy of incremental
algorithms~\cite{Gama09}: using a separate holdout set of patterns
for testing; or using Predictive Sequential (Prequential)  evaluation. The
latter has the advantage that no explicit test set has to be defined
and extracted from data. Discussion on how prequential error relates to holdout error can be found in~\cite{Gama09}, and analysis
of prequential statistics is available in~\cite{Daw84}. 

Average Prequential Error ($AE_{preq}$) is the average value of the
classification error when patterns are presented to the
current model and before the model can learn from them.
After evaluation, the pattern(s) can be used to update (train) the model. For this
reason this evaluation method is also referred to as the Interleaved Test-then-Train method.

Most papers do not take into consideration an explicit measure of 
how algorithms perform in terms of Learning Rate, Incorporation Costs or
Storage Requirements. Sometimes temporal plots of the prequential error are
used, where the time axis corresponds to the number of patterns used for
training. This measure can be labelled as $AE_{preq}(n)$, where $n$ is the time,
or pattern number, up to which the error is averaged.

The value of $AE_{preq}(n)$ can be calculated as:
\begin{equation}
\label{eq:prequentialError}
AE_{preq}(n) = \frac {1}{n} \sum_{i=1}^n L(p_i,o_i),
\end{equation}
where $p_i$ is the prediction for pattern $i$ and $o_i$ the observed value for $i$, while $n$ is the number of
patterns over which the error is measured, and $L$ a loss function. The loss function can be
defined as the 0-1 loss for achieving conventional classification:
\begin{equation}
\label{eq:prequentialErrorLossFunction}
L(p_i,o_i) = \left\{ 
\begin{array}{ll}
0,& \mbox{if } p_i=o_i\\
1,& \mbox{otherwise} %
\end{array}
\right..
\end{equation}

For finite-size datasets, algorithms are compared using this value over the
whole process ($AE_{preq}$), that is, we use Eq.~\ref{eq:prequentialError}
for a value of $n$ equal to the number of patterns in the dataset.

%When $AE_{preq}$ is used without an index $n$ it refers to
%the total accumulated error, that is $AE_{preq}(n)$ when $n$ is equal to the total
%number of patterns in a dataset.

It is also useful to observe the actual behavior of error
at every stage, in order to remove the effect of the accumulation
of the past errors. Thus, a third
measure is used, $AE^w_{preq}(n)$, that is, the average prequential error
calculated over a sliding window of $w$ (window size) patterns:
\begin{equation}
\label{eq:prequentialErrorWindow}
AE_{preq}^w(n) = \frac {1}{w} \sum_{i=n-w}^n L(p_i,o_i).
\end{equation}

Additional topics are being introduced in this field. One of these issues is
described in~\cite{ChenImb11}. When dealing with imbalanced class distributions,
algorithms may fail to take into account patterns with lower frequencies. We
consider that this aspect is also interesting in a non-stationary environment,
as class frequencies are not available from the outset and determining whether the
set is balanced or imbalanced cannot be decided in advance.

In~\cite{Shaker2015250} the authors introduce a novel measure for assessment of
IL algorithms, called Recovery Analysis. This metric describes behavior of the
algorithms in the moment of change between two stationary states, by measuring
the maximum decrease in performance and the recovery time required by each
algorithm.

Finally in~\cite{BifRea13Pitfalls}, the authors realize that most scenarios
depend on the assumption of temporal independence of data and propose a procedure and a
metric that can be used when that assumption is not true.

\subsection{Datasets}

Several artificial and real-world datasets have been used to test algorithms
with non-stationary data. Most of these datasets were produced from the Concept Drift
field. Several datasets are referenced in~\cite{bifet09}, with some of these sets 
being presented in Table~\ref{tab:oldDatasets}.

\begin{table}
\renewcommand{\arraystretch}{1.3}
\caption{Relation of datasets frequently used in literature for learning from non-stationary data. A: Artificial
Dataset, R: Real dataset, At: number of attributes, Cl: number of classes }
\label{tab:oldDatasets}
\centering
\begin{tabular}{lccccc}
\hline
Dataset                & Reference                & A/R & At  & Cl  \\ 
\hline
SEA concepts generator & \cite{StreetSEA01}       & A  & 3    & 2   \\ 
STAGGER concepts       & \cite{SchliSTAGGER86}    & A  & 3    & 2   \\ 
LED Generator          & \cite{breiman1999cart}   & A  & 24   & 2   \\
Rotating Hyperplane    & \cite{Hulten01CVFDT}     & A  & V    & 2   \\
Rotating Gaussian      & \cite{Lee10}             & A  & 2    & 2   \\
Electricity dataset    & \cite{Harries99splice}   & R  & 7    & 2   \\
%Stock market data        & \cite{LastKleKa01}      & R  & 3  & 5    \\
% Manufacturing data       & \cite{Last02}           & R  & 7  & 3     \\
% Remove ~\cite{Last02}
Traffic data streams     & \cite{Cohen2008rtiolin} & R  & 12 & 7     \\
Brain-computer interface & \cite{Lowne10}          & R  & 5  & 2    \\
Mountain Fire scenario   & \cite{Lee10}            & R  & 3  & 2    \\
\hline
\end{tabular}
%\medskip
\end{table}

Unfortunately, these datasets are being used without much information concerning the
type of problems they bring to the learning algorithms. It would be desirable
to have some information on the specific features that these datasets are
expected to test in algorithms. Also, there is a lack of reference values
to which actual performance on these datasets can be evaluated, and most papers
basically compare the proposed algorithm in some of these datasets without
further information on how the selected dataset was chosen.

For artificial datasets, another problem arises when instead of the datasets
themselves, only dataset generation algorithms are described, based on some parameters
and/or random initialization.  Parameters may include dimensionality, size of
concept drift, etc. Also, some authors use random radial basis functions 
(Random RBF generators)~\cite{bifet09}. 
In those cases it is in general impossible to determine whether actual data
sources were close, distant, separable or superimposed. Some authors describe
their artificial datasets with more detail. For instance, in~\cite{MinYao12}
several artificial datasets are proposed. For each dataset, the type of
change is described, and for each dataset a Severity measure indicates the
percentage of the input space whose class changes after the drift
is complete.

\section{IL Evaluation Methodology}

In this work we propose an evaluation methodology for testing the capabilities
of any algorithm in dealing with incremental learning tasks. The aim of this methodology is
to formalize a test bench that can be used to evaluate incremental algorithms able to learn from non-stationary data. 

In this way we design a set of evaluation tests, corresponding to features
that should be considered in the evaluation of any incremental learning algorithm. Each
evaluation test is composed of a dataset, its optimal Bayes error (that is,
a minimal error bound for the dataset) and the description of the dynamics of
the data transformation, including changes in class distributions and class priors.

\subsection{Algorithms}
\label{sec:algorithms}

Before presenting the various test benchmarks, which is the core contribution of this paper, 
we want to introduce the various learning methods that will be tested against these benchmarks,
These methods will be tested to assert that the proposed benchmarks are properly measuring 
the specific property that we want to assess.

This selection of reference algorithms is by no means exhaustive over all possible 
incremental algorithms in the literature. We selected a varied set of algorithms 
whose behaviors, strengths and weaknesses are well known and understood.
Indeed, by using algorithms of different types, we hope to see how different mechanisms
are responsible for good or bad results in each of the defined objectives.

We have used the Massive Online Analysis (MOA) \cite{MOA10} software 
environment to provide the algorithms and run the experiments. 
Among the algorithms it provides, we have selected the following:

Naive Nayes (NB) constructs a static model from which we can assess how
classifiers designed for stationary data degrade in the non-stationary case. This version of Naive Bayes
calculates its model incrementally.

Stochastic Gradient Descent (SGD) has been used to train classifiers for
large-scale problems (e.g.
~\cite{bottou10}). We use it for NSL because, under certain assumptions
related to the number and type of classifier, it is able to quickly adapt to non-stationary data. 
The version we selected uses a two-class SVM with linear kernel. 

Dynamic Weighted Majority (DWM)~\cite{Kolter07} is an ensemble method that is
able to classify non-stationary data using a dynamic weight assignment to base classifiers that depends
on their current accuracy. Classifiers are created or pruned dynamically.
We used a Naive Bayes base learner for DWM.

OZABAG-ADWIN~\cite{bifet09} is an online bagging method that is adapted to
concept drift by using a self-adapting window of patterns for training.
Window size is changed depending on a decision function called ADWIN. This type
of change detector monitors a random variable and decides if a change has
taken place depending on the result of a statistical test performed on the value of
the variable measured over two different sets of data: past data and current
data. For instance, ADWIN detects change in the mean values of the false
positive rate of the classifier. If both values of the averages false positive rate differ
significantly given a certain confidence value, the window size is reduced by
removing old data. We used a ensemble of $10$ base classifiers, each one
learns a Hoeffding tree structure using CVFDT~\cite{Hulten01CVFDT}.

Although NN classification may seem elementary, we use this approach to assess
the success of time-based forgetfulness. By using a fixed-size sliding window,
the algorithm will use only recent data for classification. Obviously
the selection of the proper window size becomes the problem in that case.
We therefore investigate three window sizes: NN$^{100}$, NN$^{1500}$, 
and NN$^{6000}$. The smaller window (100) tends to favor adaptation to change, while the larger 
one (6000) tends to increase accuracy.

We must remark here that it is not our intention, per se, to test the accuracy and performance
of these methods. The methods were not chosen for their particular performance or popularity, 
but rather to set up an experimental environment to validate the efficacy of our methodology 
for the comparison of different incremental learning methods.

\subsection{Properties of interest}
\label{sec:properties}

The key point, when designing a testing methodology, is to determine exactly what needs to be measured. For instance, we might wish to evaluate ``plasticity'', but this type of concept is too vague to be practical. Another approach would be to enumerate all possible data transformations. Assuming Gaussian distributions, this means moving centers and changing variances in different directions, using different dynamics. Obviously, we need to restrict ourselves to a subset of possibilities.

In this context, the following properties were selected in order to guide the creation of data models that are apt at evaluating the adaptive capabilities of incremental learning algorithms:
\begin{enumerate}
\item Contradiction in the presence of global change; contradiction refers to non stationarity that causes new knowledge to disagree with older knowledge; adaptability to this requires some form of forgetfulness; global change refers to distributions that simultaneously all undergo similar transformations.
\item Contradiction in the presence of local change; in contrast to global change, local change refers to different transformations applied to different distributions.
\item Long-term memory; refers to past knowledge that has not been used for a while, but then becomes useful again; adaptability to this assumes that forgetting mechanisms are not exclusively time-based.
\item Change in distribution frequency; refers to priors that change over time.
\item Appearance of new concepts; refers to new distributions that are, at some point in time, added to existing mixtures, or that create whole new classes.
\item Speed of change; refers to the scale of non-stationarity, that is slow vs fast changes.
\item Effect of dimensionality; refers to the input feature space dimensionality.
\item Complex non-stationary changes; refers to combinations of other relevant properties.
\end{enumerate}

\subsection{Data Modeling and Generation}
In the following, we model data using a mixture of mixtures. Let $C_i^t$ denote the $i^{th}$ class at time $t$, $i=1,\ldots,K^t$, of a non-stationary $K^t$-class problem. Each class is modeled by a mixture of $k_i^t$ distributions $\{{\cal G}_{i,1}^t, \ldots, {\cal G}_{i,k_i^t}^t\}$, where exponent $t$ stresses the time dependence of the distributions. Then, the density function $p(\mathbf{x}, t)$ represents the probability of observing vector $\mathbf x$ at time $t$:
\begin{equation}
p(\mathbf{x}, t) = \sum_{i=1}^{K^t} \sum_{j=1}^{k_i^t} P({\cal G}_{i,j}^t)\,p(\mathbf{x}\,|\,{\cal G}_{i,j}^t),
\end{equation}
where $P({\cal G}_{i,j}^t)$ is the a priori probability of the $j^{th}$ distribution of $C_i^t$ and $p(\mathbf{x}\,|\,{\cal G}_{i,j}^t)$ the density function of ${\cal G}_{i,j}^t$. With such a model, the probability that $\mathbf{x}$ belongs to $C_i^t$ is expressed by:
\begin{equation}
P(C_i^t\,|\,\mathbf{x},t) = \frac{\sum_{j=1}^{k_i^t} P({\cal G}_{i,j}^t)\,p(\mathbf{x}\,|\,{\cal G}_{i,j}^t)}{p(\mathbf{x}, t)}.
\label{eq:groundtruth}
\end{equation}
This equation defines the ground truth. The optimal Bayesian classifier is 
obtained by selecting, for a pattern
$\mathbf{x}$, the class $i$ that maximices $P(C_i^t\,|\,\mathbf{x},t)$.

If we arbitrarily set $p(\mathbf{x}\,|\,{\cal G}_{i,j}^t)\sim {\cal
N}(\bm{\mu}_{i,j}^t, \bm{\Sigma}_{i,j}^t)$, a multidimensional Gaussian distribution with center $\bm{\mu}_{i,j}^t$ and covariance matrix $\bm{\Sigma}_{i,j}^t$, the data model becomes a mixture of Gaussian mixtures.

To generate data using this model, at each discrete time step $t$, we first randomly select distribution ${\cal G}_{s}^t$ with probability:
\begin{equation}
P_s = \frac{P({\cal G}_{s}^t)}{\sum_{i=1}^{K^t} \sum_{j=1}^{k_i^t} P({\cal G}_{i,j}^t)},
\label{eq:roulette}
\end{equation}
using a proportional random selection, and then generate $\mathbf{x}$ using $p(\mathbf{x}\,|\,{\cal G}_{s}^t)$.

\subsection{Model Description}

Models are described by a set of named classes (typically ``A'', ``B'', ``C'', etc.), composed of a weight parameter plus a mixture of Gaussian distributions. Distributions are specified by a start time index, a weight parameter, an initial position (center), an initial covariance matrix, and a sequence of cascading transforms. Each transform is associated with a duration parameter, during which a shape preserving (similarity) transformation is applied linearly; the transform is defined by rotation angles, scale factor, and translation.

The start index of a distribution determines at which time it begins to exist. Before that index, it is not part of the class mixture. The class itself starts to exist at the beginning of its first distribution. The duration of a distribution is equal to the sum of the durations of its cascade of transforms. To exist, a distribution must thus define at least one transform with a non null duration. To set a stationary distribution, an identity transform should be specified (the default), that is a transform with unit scale, null rotation, and null translation.

The rotation and scale parameters of a transform are applied to the covariance matrix of its Gaussian distribution, while the translation parameters are applied to its center position. The transforms are sequential and cumulative, with each transform defining a separate phase of non-stationarity. The distribution at the beginning of a new transform is equal to the distribution at the end of the previous transform. A transform is applied gradually, in a linear fashion, during its full duration. For an instantaneous transformation, the duration parameter should be set to zero.

The weights of classes and distributions are used to determine the distribution priors that enable data generation. Let $w_i$ denote the weight associated with class $C_i$, and $w_{i,j}$ the weight of distribution ${\cal G}_{i,j}$. Then, prior $P({\cal G}_{s,v}^t)$ of some active distribution ${\cal G}_{s,v}^t$ at time $t$ is computed by:
\begin{equation}
P({\cal G}_{s,v}^t) = \frac{w_s w_{s,v}}{\sum_{i=1}^{K^t}w_i\sum_{j=1}^{k_i^t}w_{i,j}},
\end{equation}
where $K^t$ is the number of active classes at time $t$, and $k_i^t$ the number of active distributions within the mixture of class $C_i^t$.

\section{Datasets and results}
\label{sec:datasets}

In this section, we go through all of the properties that were presented in Section~\ref{sec:properties} and describe the corresponding datasets that are proposed to test these properties. We then apply these datasets to all of the algorithms presented in Section~\ref{sec:algorithms} and report results.

\begin{table}
\centering
\caption{Results for all datasets, values are final $AE_{preq}$ in \%.} 
\label{tab:summary}
\resizebox{0.475\textwidth}{!}{
\begin{tabular}{lrrrrrrrr}
  \hline
Problem & Opt. & NB & SGD & DWM & OZAB & NN$^{100}$ & NN$^{1500}$ & NN$^{6000}$ \\ 
  \hline
  NSGT & \textit{2.95} & 25.27 & 7.68 & \textbf{4.21} & \textbf{4.20} & 4.83 & 6.77 & 10.97 \\ 
  NSGT-F & \textit{2.91} & 41.73 & 14.14 & 6.90 & \textbf{4.39} & 4.77 & 11.79 & 12.04 \\ 
  NSGR & \textit{0.00} & 49.61 & 0.04 & 0.90 & 0.54 & \textbf{0.02} & \textbf{0.02} & 36.95 \\ 
  NSLC & \textit{4.05} & 6.44 & \textbf{4.25} & 4.91 & 4.72 & 6.19 & 6.12 & 7.77 \\ 
  NSGT-I & \textit{2.93} & 25.05 & 8.02 & 5.85 & \textbf{4.48} & 4.80 & 9.58 & 10.28 \\ 
  NSPC & \textit{5.76} & \textbf{5.94} & 6.77 & 6.24 & 6.16 & 9.00 & 8.73 & 8.90 \\ 
  NSPC-A & \textit{5.37} & 6.09 & \textbf{5.89} & 6.09 & \textbf{5.98} & 8.45 & 8.28 & 8.82 \\ 
  NSGT-5D & \textit{5.74} & 25.43 & 9.35 & \textbf{7.34} & 7.63 & 12.18 & 11.16 & 11.88 \\ 
  NSCX & \textit{4.18} & 14.28 & 12.94 & 6.90 & \textbf{6.08} & 6.47 & 8.62 & 10.19 \\ 
   \hline
\end{tabular}
}
\end{table}

In Table~\ref{tab:datasets-summary} we summarize the parameters used for each of
the datasets. All datasets are two-dimensional and have $10000$ patterns.
For each distribution in a dataset, a sequence of transitions describe the changes 
applied to the distribution parameters. Distributions and transitions are also
described graphically in Fig~\ref{fig:AllDatasets:Plot}.

%\onecolumn
%\begin{sidewaystable}
\begin{table*}
\caption{Summary of dataset parameters. Transform types:
rmoveto$(r_x,r_y)$, relative translation of $r_x,r_y$ units; wchangeto$(w)$,
linear change from current weight to $w$; scale$(s)$, linear change in variance
until its value is multiplied by s; rotate$(g)$, rotation of $g$ degrees}
\label{tab:datasets-summary}
\centering
%\medskip
%\small
%\renewcommand{\arraystretch}{1.2}
\resizebox{\textwidth}{!}{
\begin{tabular}{|c|c|c|c|c|c|c|c|c|p{5cm}|}
\hline
\multirow{2}{*}{dataset} & \multirow{2}{*}{dims} & \multirow{2}{*}{size} & \multicolumn{5}{c|}{initial distributions parameters} & \multicolumn{2}{c|}{phases of transformation}\\
\cline{4-10}
 & & & class & weight & center & stddev & rotation & period & transform type\\
\hline\hline
\multirow{2}{*}{NSGT} & \multirow{2}{*}{2} & \multirow{2}{*}{10001} & A &
\multirow{2}{*}{1.0} & $(0.0, 0.0)$ & \multirow{2}{*}{$(2.5, 1.0)$} & $45^\circ$
& \multirow{2}{*}{0-9999} & \multirow{2}{*}{rmoveto$(10,10)$}\\
 & & & B & & $(5.0,0.0)$ & & $-45^\circ$ &  & \\
 \hline\hline
% \multirow{2}{*}{NSGT-F} & \multirow{2}{*}{2} & \multirow{2}{*}{10001} & A &
% \multirow{2}{*}{0.5} & $(0.0, 0.0)$ & \multirow{2}{*}{$(2.5, 1.0)$} & $45^\circ$
% & \multirow{2}{*}{0-9999} & \multirow{2}{*}{rmoveto$(30,30)$}\\
%  & & & B & & $(5.0,0.0)$ & & $-45^\circ$ &  & \\
%  \hline\hline
NSGT-F & \multicolumn{7}{c|} {Same as NSGT} & 0-9999 & rmoveto$(30,30)$\\
 \hline\hline
\multirow{2}{*}{NSGR} & \multirow{2}{*}{2} & \multirow{2}{*}{10001} & A &
\multirow{2}{*}{1.0} & $(10.0,0.0)$ & \multirow{2}{*}{($2.0, 5.0$)} &
\multirow{2}{*}{$45^\circ$} & \multirow{2}{*}{0-9999} & \multirow{2}{*}{rotate
once around origin}\\
 & & & B & & $(-10.0,0.0)$ & & & & \\
\hline\hline
\multirow{2}{*}{NSLC} & \multirow{2}{*}{2} & \multirow{2}{*}{10001} & A &
\multirow{2}{*}{1.0} & $(-2.0,2.0)$ & \multirow{2}{*}{$(2.5, 1.0)$} & $45^\circ$
& \multirow{2}{*}{0-9999} & rmoveto$(0.0,-4.0)$\\
 & & & B & & $(2.0,-2.0)$ & & $-45^\circ$ &  & rmoveto$(0.0,4.0)$\\
\hline\hline
\multirow{6}{*}{NSGT-I} & \multirow{6}{*}{2} & \multirow{6}{*}{10001} & A &
\multirow{6}{*}{1.0} & $(0.0, 0.0)$ & \multirow{6}{*}{$(2.5, 1.0)$} &
\multirow{3}{*}{$45^\circ$} & 0-4999 & rmoveto{$(10,10)$}\\
 &  & &   & & &  &  & 5000 & rmoveto{$(-10,-10)$}\\
 &  & &   & & &  &  & 5001-10000 & rmoveto{$(10,10)$}\\
%  \cline{4-10}
 &  & & B & & $(5.0,0.0)$ & & \multirow{3}{*}{$-45^\circ$} & 0-4999 & rmoveto{$(10,10)$} \\
 &  & &   & & &  &  & 5000 & rmoveto{$(-10,-10)$}\\
 &  & &   & & &  &  & 5001-10000 & rmoveto{$(10,10)$}\\
\hline\hline

%\multirow{4}{*}{NSGT-I} & \multirow{4}{*}{2} & \multirow{4}{*}{10001} & A$_1$ &
%\multirow{2}{*}{0.5} & $(0.0, 0.0)$ & \multirow{2}{*}{$(2.5, 1.0)$} &
% $45^\circ$ & \multirow{2}{*}{0-4999} & \multirow{2}{*}{rmoveto$(10,10)$}\\
% & & & B$_1$ & & $(5.0,0.0)$ & & $-45^\circ$ &  & \\
%  \cline{4-10}
%                         &                      &                      &  
%                         % A$_2$ & \multirow{2}{*}{0.5} & $(0.0, 0.0)$ &
                         % \multirow{2}{*}{$(2.5, 1.0)$} & $45^\circ$ &
                         % \multirow{2}{*}{5000-9999} & \multirow{2}{*}{rmoveto$(10,10)$}\\
% & & & B$_2$ & & $(5.0,0.0)$ & & $-45^\circ$ &  & \\
%\hline\hline
\multirow{3}{*}{NSPC} & \multirow{3}{*}{2} & \multirow{3}{*}{10001} & A$_1$ &
$0.05$ & $(-2.0,0.0)$ & \multirow{2}{*}{$(2.5, 1.0)$} & $45^\circ$
& \multirow{2}{*}{500-9499} & wchangeto$(0.45)$ \\
 & & & A$_2$ & $0.45$ & $(2.0,0.0)$ & & $-45^\circ$ &  & wchangeto$(0.05)$ \\
  \cline{4-10}
 & & & B & $0.5$ & $(0.0,3.5)$ & $(1.0, 1.0)$  & $0^\circ$ & 
 \multicolumn{2}{c|}{}  \\
 \hline\hline
% NSPC-F & \multirow{3}{*}{2} & \multirow{3}{*}{10001} & A$_1$
% & \multicolumn{4}{c|} {Same as NSPC} &
% \multirow{2}{*}{4500-5499} & wchangeto$(0.45)$  \\
%        &  &  & A$_2$ &  \multicolumn{4}{c|} {} &   & 
%        wchangeto$(0.05)$  \\
%   \cline{4-10}
%        &  &  & B & \multicolumn{4}{c|} {Same as NSPC} &   &
%          \\
%  \hline\hline
NSPC-A & \multirow{3}{*}{2} & \multirow{3}{*}{10001} & A$_1$
& 0.0 & \multicolumn{3}{c|} {Same as NSPC} &
\multirow{2}{*}{5000} & wchangeto$(0.5)$  \\
       &  &  & A$_2$ & 0.5 & \multicolumn{3}{c|} {} &   & 
       wchangeto$(0.0)$  \\
  \cline{4-10}
       &  &  & B & 0.5 & \multicolumn{3}{c|} {Same as NSPC} &
       \multicolumn{2}{c|}{}          \\
 \hline\hline
\multirow{2}{*}{NSGT-5D} & \multirow{2}{*}{5} & \multirow{2}{*}{10001} & A &
\multirow{2}{*}{0.5} & $(0.0, 0.0, 0.0, 0.0, 0.0)$ & $(1.0,
1.0, 1.0, $  & \multirow{2}{*}{$0^\circ$} & \multirow{2}{*}{0-9999} &
\multirow{2}{*}{rmoveto$(6.3, 6.3, 6.3, 6.3, 6.3)$}\\
 & & & B & & $(3.15, 0.0, 0.0, 0.0, 0.0)$ & $1.0, 1.0)$ &  &  & \\
\hline\hline
%%%% NSCX
\multirow{7}{*}{NSCX} & \multirow{7}{*}{2} & \multirow{7}{*}{10001} &
\multirow{2}{*}{A$_1$} &  \multirow{2}{*}{$0.65$} & 
\multirow{2}{*}{$(0.0,0.0)$} &  \multirow{2}{*}{$(2.5, 1.0)$} & 
\multirow{2}{*}{$30^\circ$} & 0-4999 & rmoveto$(5.0,5.0)$, rotate$(90^\circ)$,
scale$(2)$
\\
            &  &  &  & &  &  & & 5000-9999 & rmoveto$(5.0,5.0)$,
            wchangeto$(1.00)$
            \\
            & & & A$_2$ & $0.35$ & $(0.0,-4.0)$ & $(0.6, 2.0)$ & $0^\circ$ &
             &  \\
  \cline{4-10}
 & & & \multirow{7}{*}{B} & $0.0$ & \multirow{7}{*}{$(-2.0,3.0)$} &
 \multirow{7}{*}{$(1.5, 0.5)$} & \multirow{7}{*}{$0^\circ$} & 0-499 &
 wchangeto$(0.2)$ \\
 & & &   & $0.2$ &  &   &  &
 500-1999 & rmoveto$(3.0,-4.0)$,rotate$(30^\circ)$,
 wchangeto$(0.5)$ \\
 & & &   & $0.5$ &  &   &  & 2000-4499 &
 rmoveto$(4.0,-1.0)$,rotate$(30^\circ)$,
 wchangeto$(0.8)$ \\
 & & &   & $0.8$ &  &   &  & 4500-9999 &
 rmoveto$(6.0,5.0)$,rotate$(30^\circ)$,
 wchangeto$(1.0)$ \\
\hline
\end{tabular}
}
\end{table*} 
%\end{sidewaystable}
%\twocolumn

We have generated ten versions of each of the datasets,
using different random seeds for sampling. Hereafter we report average results for these ten runs, and
graphics are also generated using the average result for each data point.
This allows a reduced variance in the average results and produces smoother
graphics. Whenever results are shown, \textbf{boldface} indicates the best
algorithm. This comparison does not take into account the Bayes Optimal
classifier which is listed in the first column. All algorithms whose
average result is not significantly different from the best,
with a significance value of $\alpha=0.05$, are also in \textbf{boldface}. This
comparison was made using pairwise Wilcoxon tests.

Results of these experiments are summarized in Table~\ref{tab:summary}, where
we show the value of the Average Prequential Error ($AE_{preq}$, see 
Eq.~\ref{eq:prequentialError}) at the end of the datasets.

\subsection{Contradiction in the Presence of Global Change}
\label{sec:ContradictionGlobalChange}

The first two datasets seek to evaluate the adaptability to global
contradiction, that is, the ability of algorithms to modify their internal model
in order to globally follow moving distributions that are linked together in some way. In these datasets, contradictions are introduced gradually, by translating both distribution centers in order to maintain their relative positions, so that the samples of one distribution gradually occupy the position of previous samples of the other.

\begin{figure*}
\centering
\subfloat[NSGT, NSGT-I and NSGT-F]{
\includegraphics[width=0.25\linewidth]{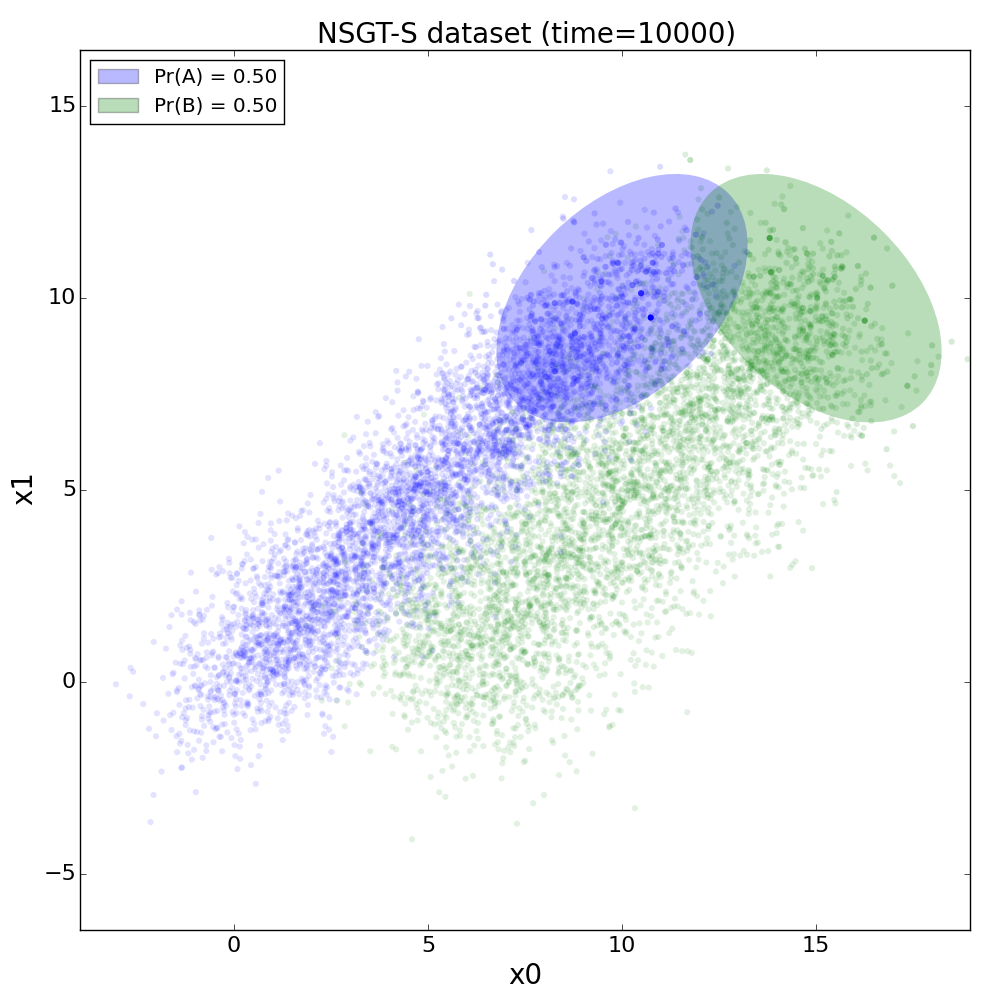}
\label{fig:NSGT:Plot}
} 
\hfil
\subfloat[NSGR]{
\includegraphics[width=0.25\linewidth]{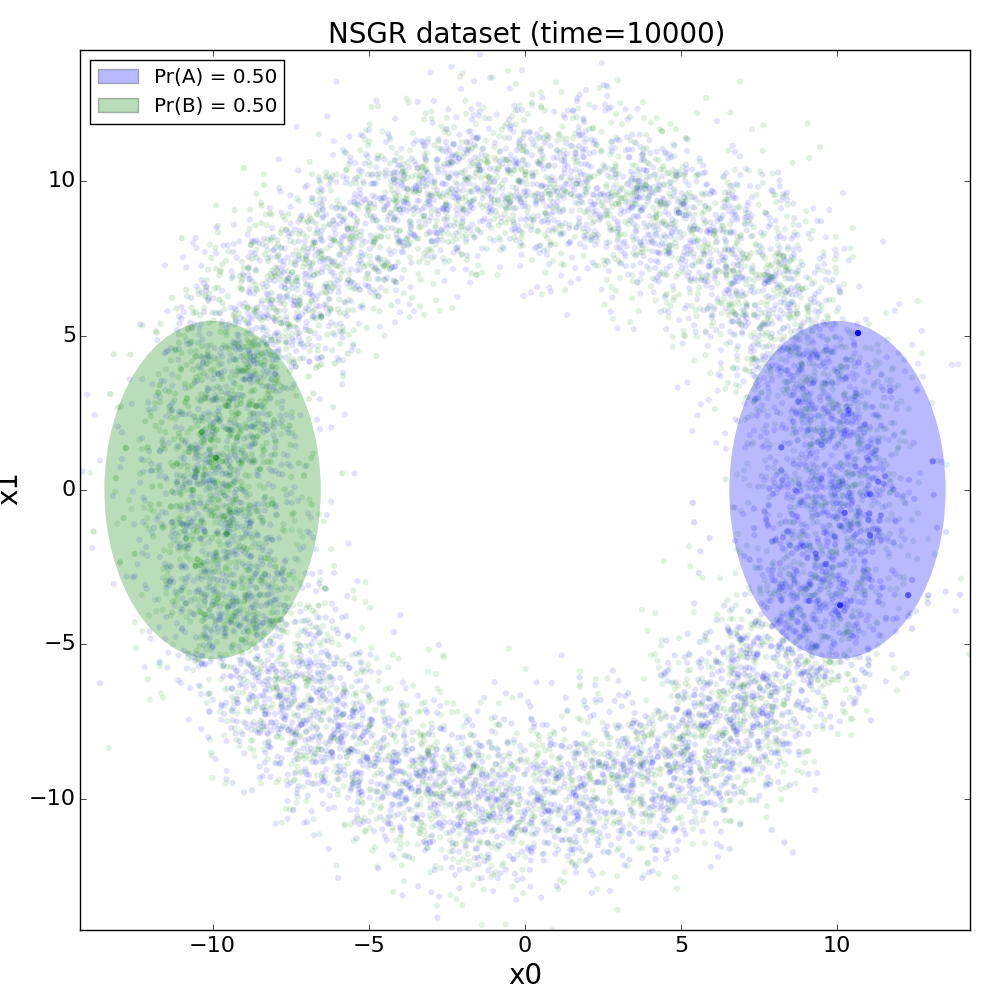}
\label{fig:NSGR:Plot}
}
\hfil
\subfloat[NSLC]{
\includegraphics[width=0.25\linewidth]{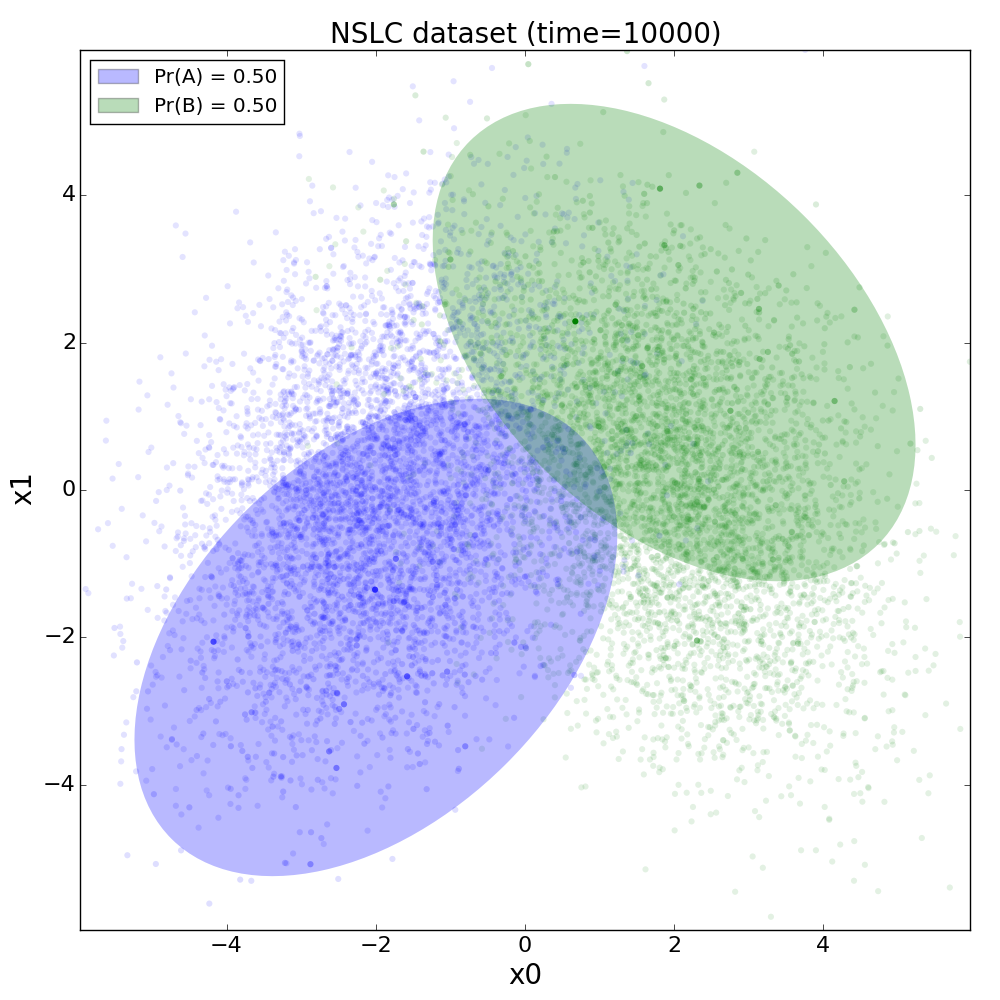}
\label{fig:NSLC:Plot}
}
\\
\subfloat[NSPC: Start]{
\includegraphics[width=0.25\textwidth]{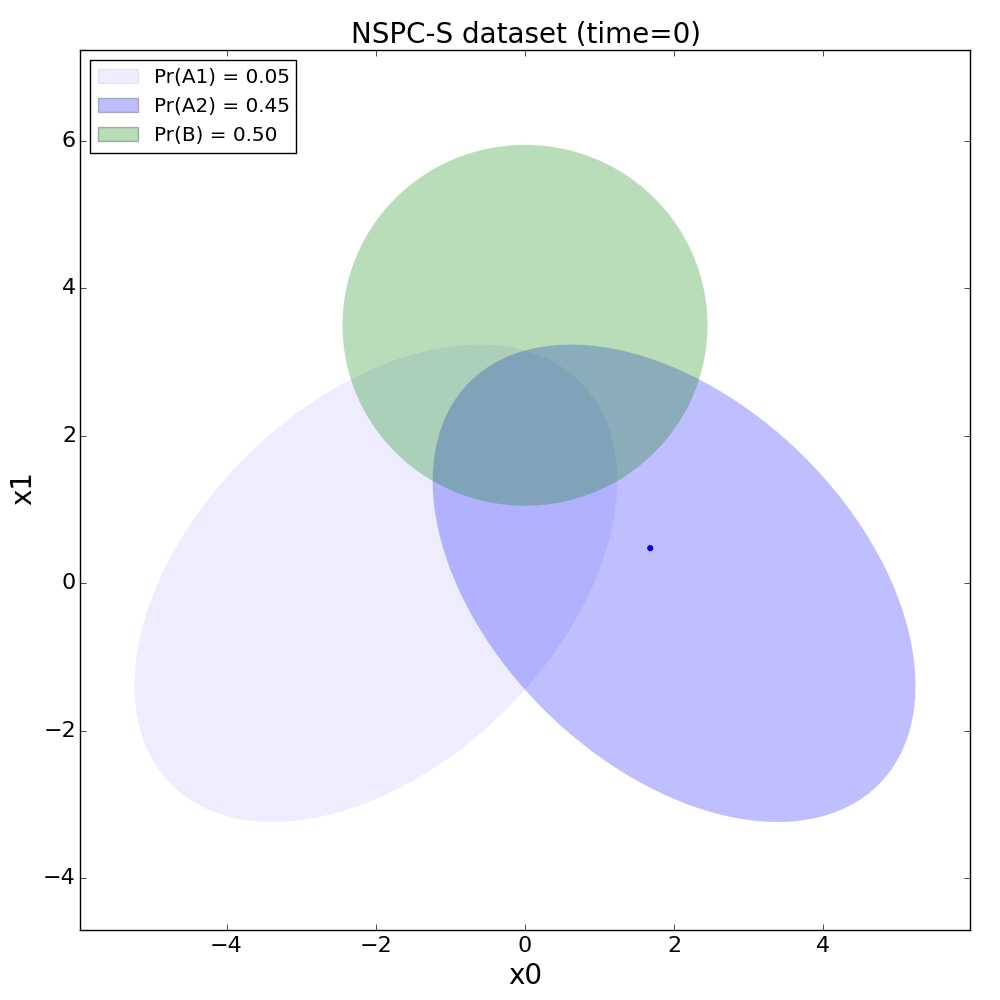}
\label{fig:NSPC-S:Plot:Start}
}
\hfil
\subfloat[NSPC: Pattern 5000]{
 \includegraphics[width=0.25\textwidth]{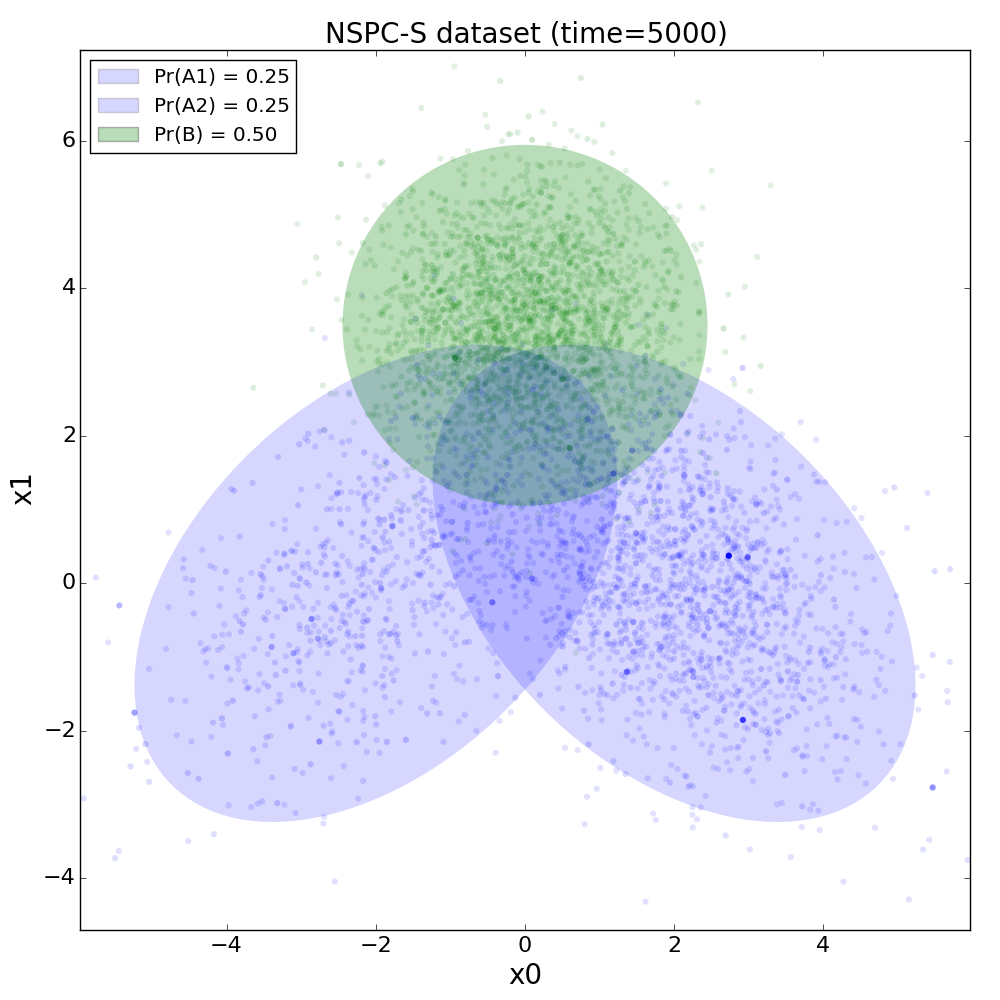}
 \label{fig:NSPC-S:Plot:5000}
 }
\hfil
\subfloat[NSPC: Pattern 10000]{
\includegraphics[width=0.25\textwidth]{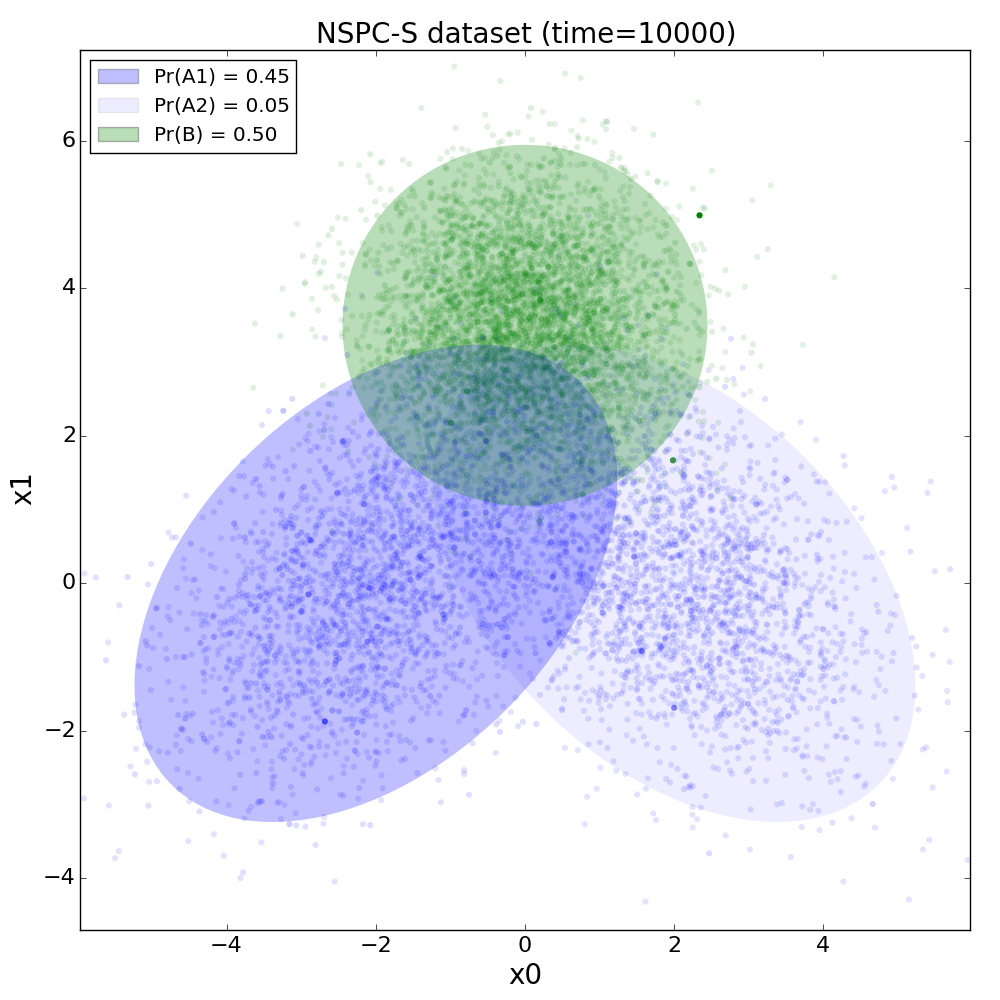}
\label{fig:NSPC-S:Plot:End}
}
\\
\subfloat[NSCX: Start]{
\includegraphics[width=0.25\textwidth]{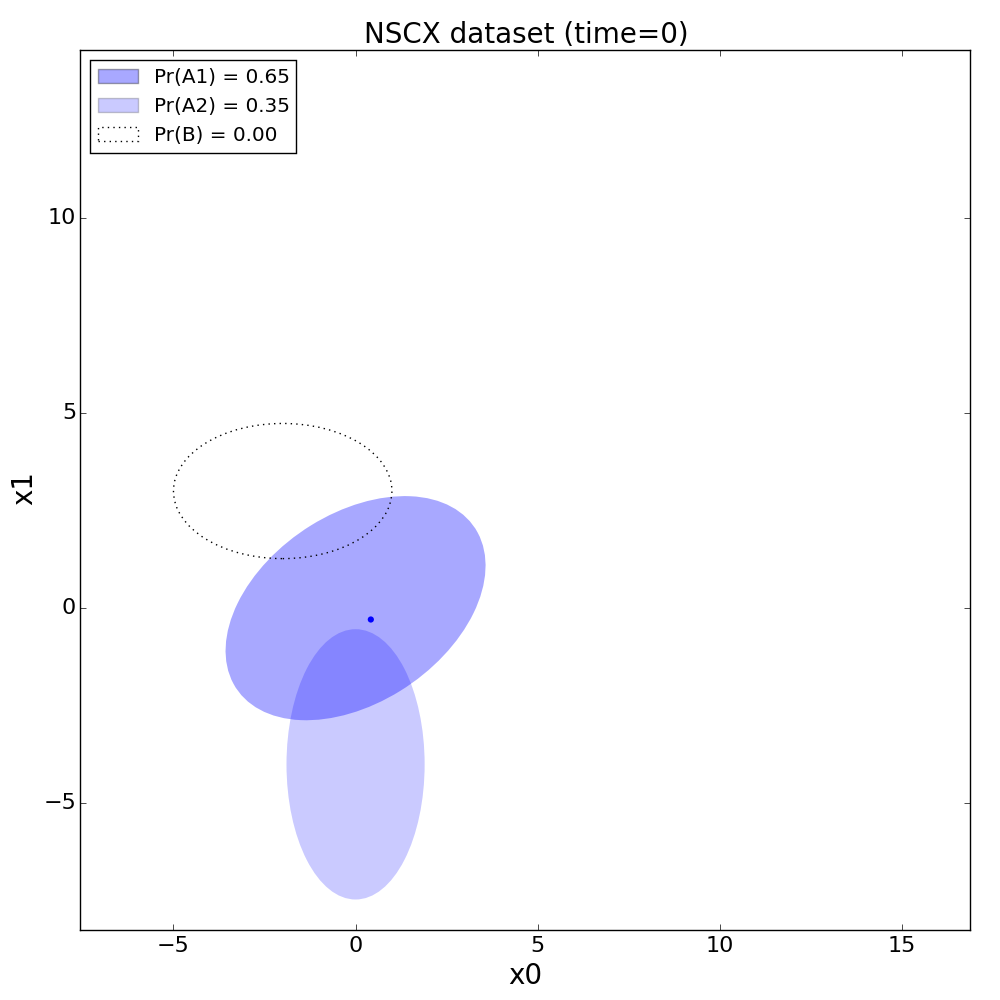}
\label{fig:NSCX:Plot:Start}
}
\hfil
\subfloat[NSCX: Pattern 5000]{
\includegraphics[width=0.25\textwidth]{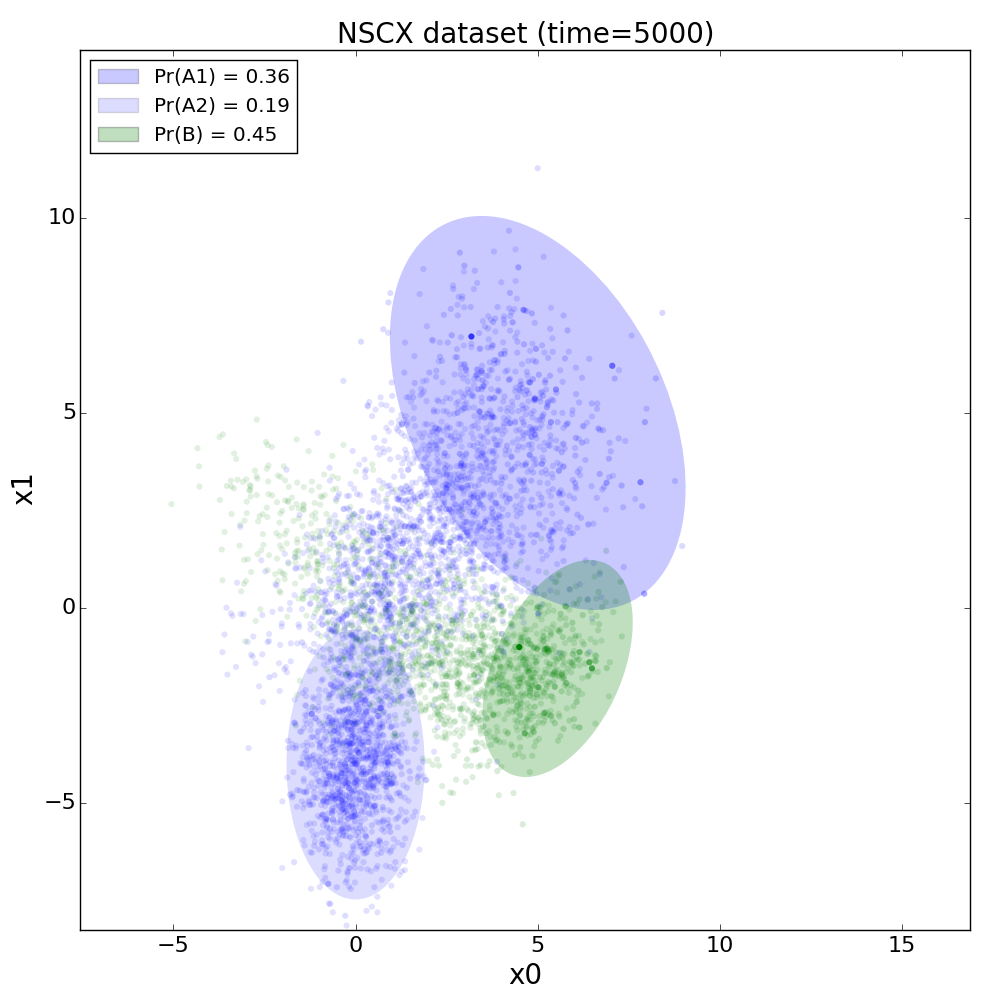}
\label{fig:NSCX:Plot:5000}
}
\hfil
\subfloat[NSCX: Pattern 10000]{
\includegraphics[width=0.25\textwidth]{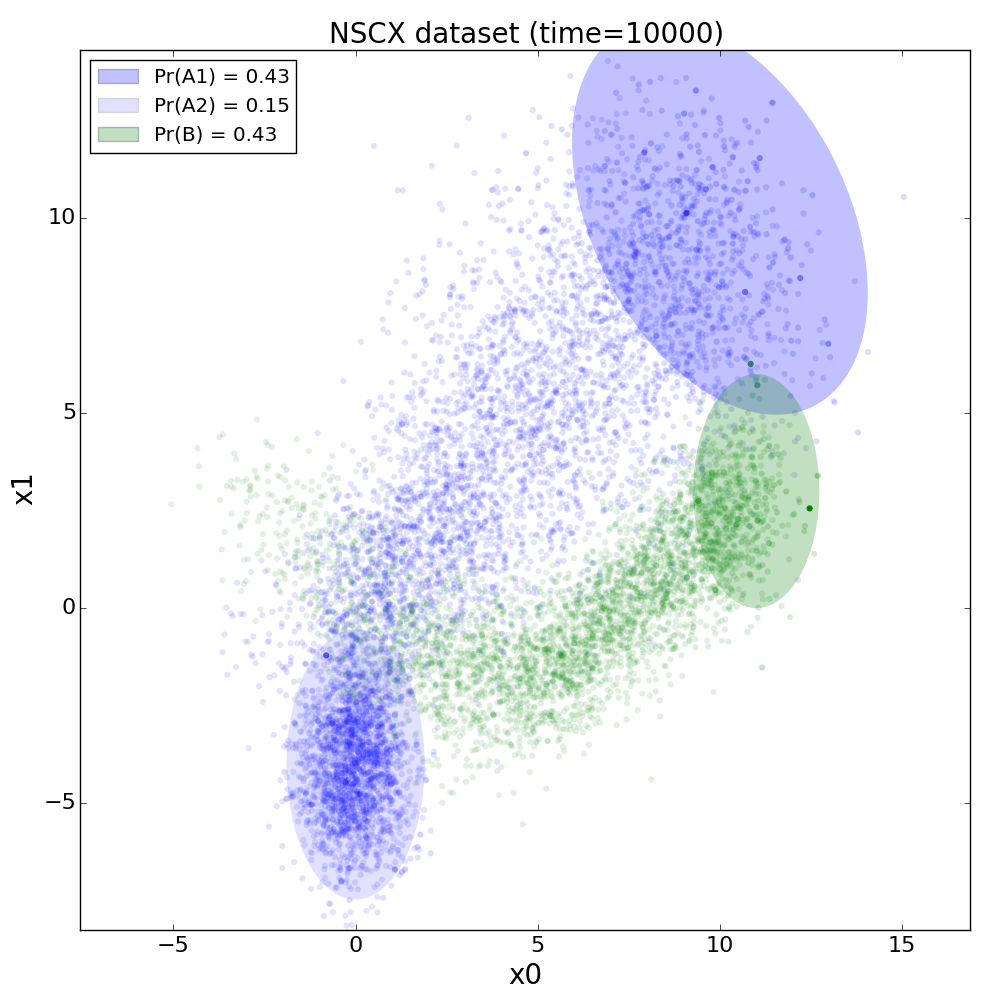}
\label{fig:NSCX:Plot:End}
}
\caption{Graphical description for all the datasets. Blue color is distribution
A (or A1 and A2 in some cases) and green is B (B1 and B2). Ellipses are used
to represent the distributions at a given point of time, where size is a measure
of variance of the distribution. Color intensity is a measure of the \textit{a
priori} probability of the distribution. Patterns are plotted to show the
trajectories.}
\label{fig:AllDatasets:Plot}
\end{figure*}

NSGT (for non-stationary global translation) starts with two overlapping
distributions that drift linearly in a fixed direction, preserving their
relative positions and orientations, see
Fig.~\ref{fig:NSGT:Plot}. NSGR simulates the rotation of the two distribution centers around a mid-point,
see Fig.~\ref{fig:NSGR:Plot}. Contradictions will begin to appear when the angle of rotation approaches 180
degrees and will continue for an additional 180 degrees.

In Table~\ref{tab:summary} we can see
that DWN and OZAB perform equally well, while NN has a good result if we select
the smaller window size (NN$^{100}$). Thus, this dataset requires either
management of contradiction, or a short time-based memory window. SGD is able to adapt a separation surface to
some extent, however, its final accuracy is intermediate. NB is the worst, as
expected, as it is not designed to adapt to contradiction in any way.

\begin{figure*}
\centering
\subfloat[NSGT]{
\includegraphics[width=0.32\linewidth]{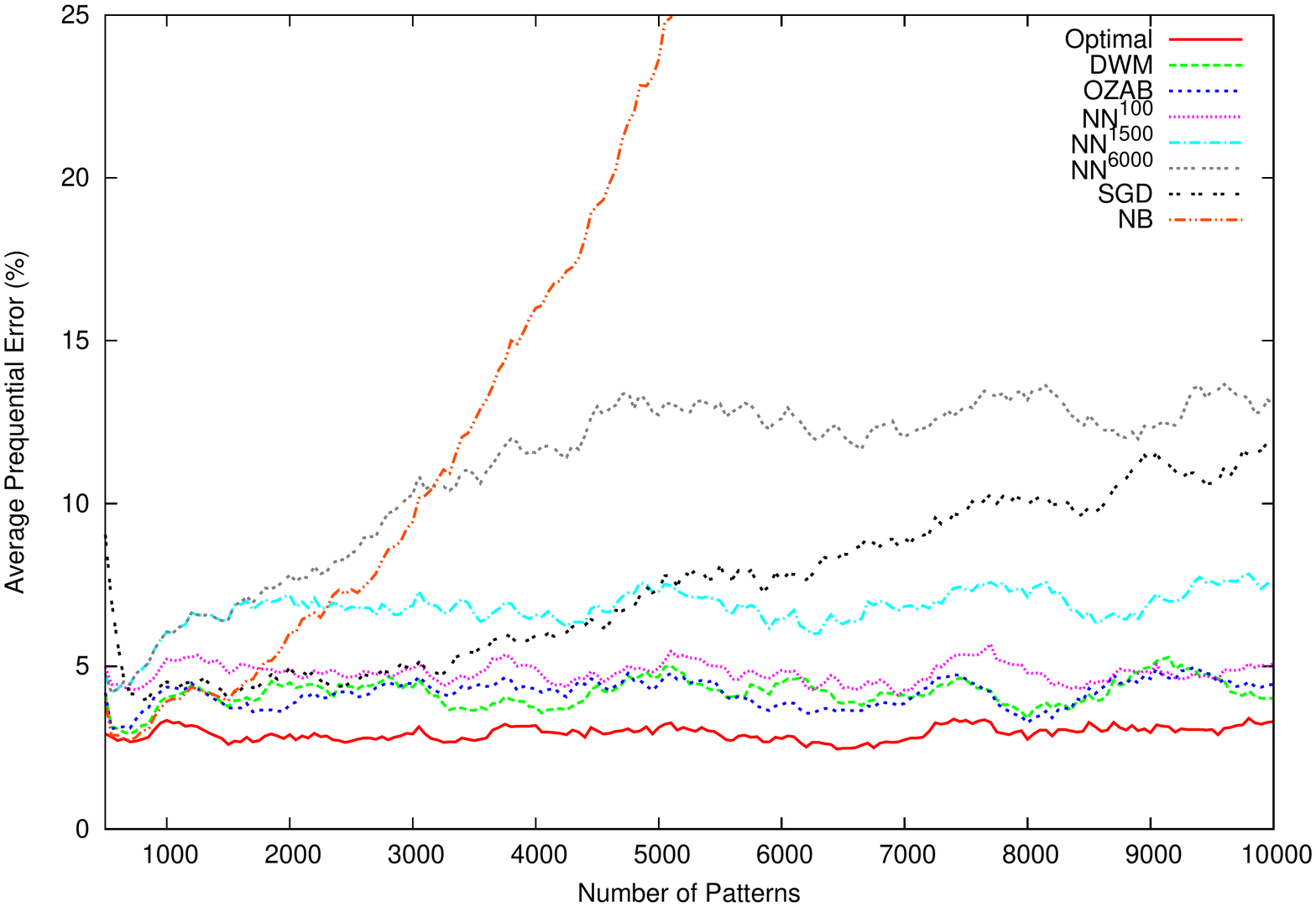}
\label{fig:NSGT-S:AE500}
}
\hfil
\subfloat[NSGR]{
\includegraphics[width=0.32\linewidth]{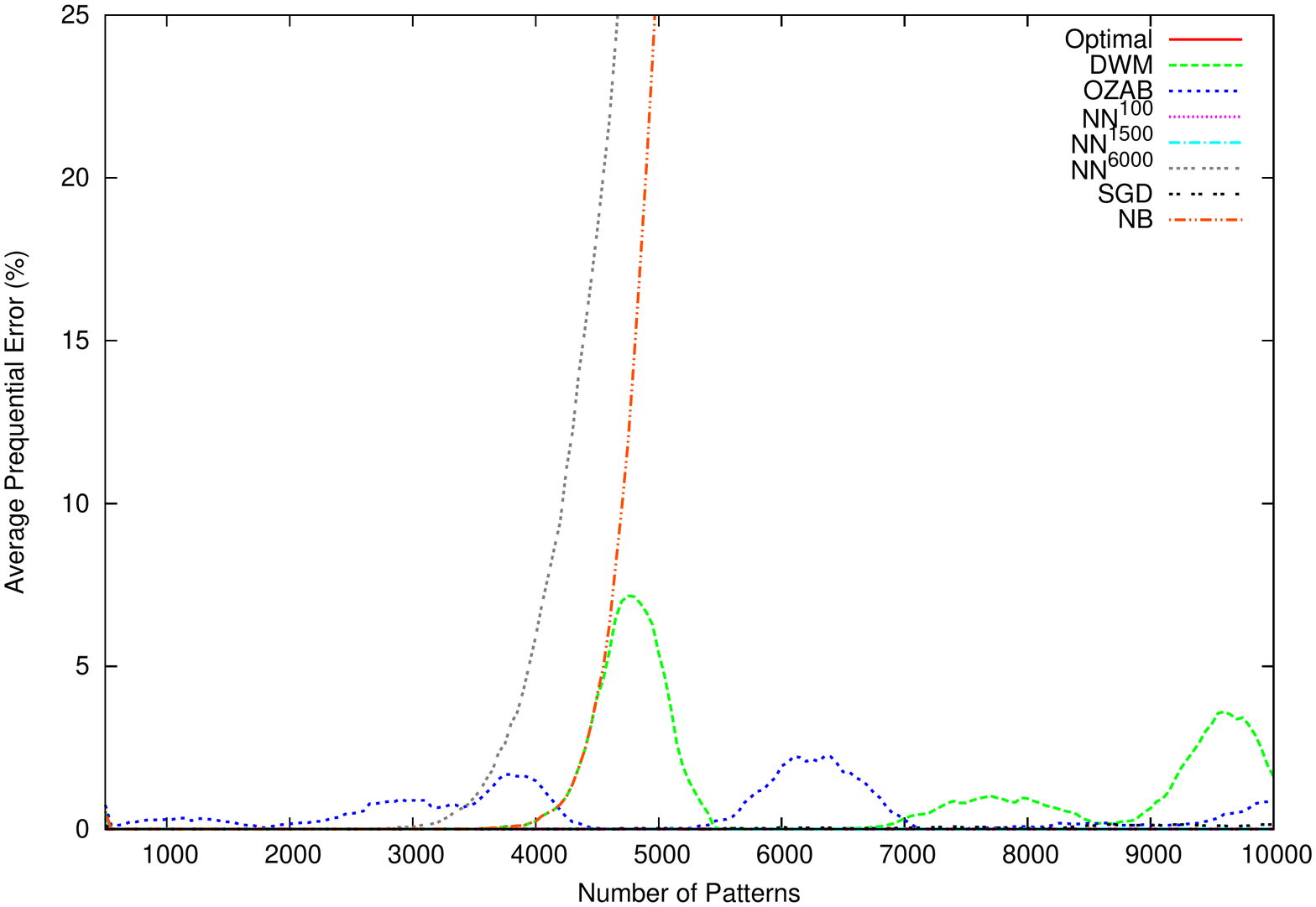}
\label{fig:NSGR:AE500}
}
\hfil
\subfloat[NSLC]{
\includegraphics[width=0.32\linewidth]{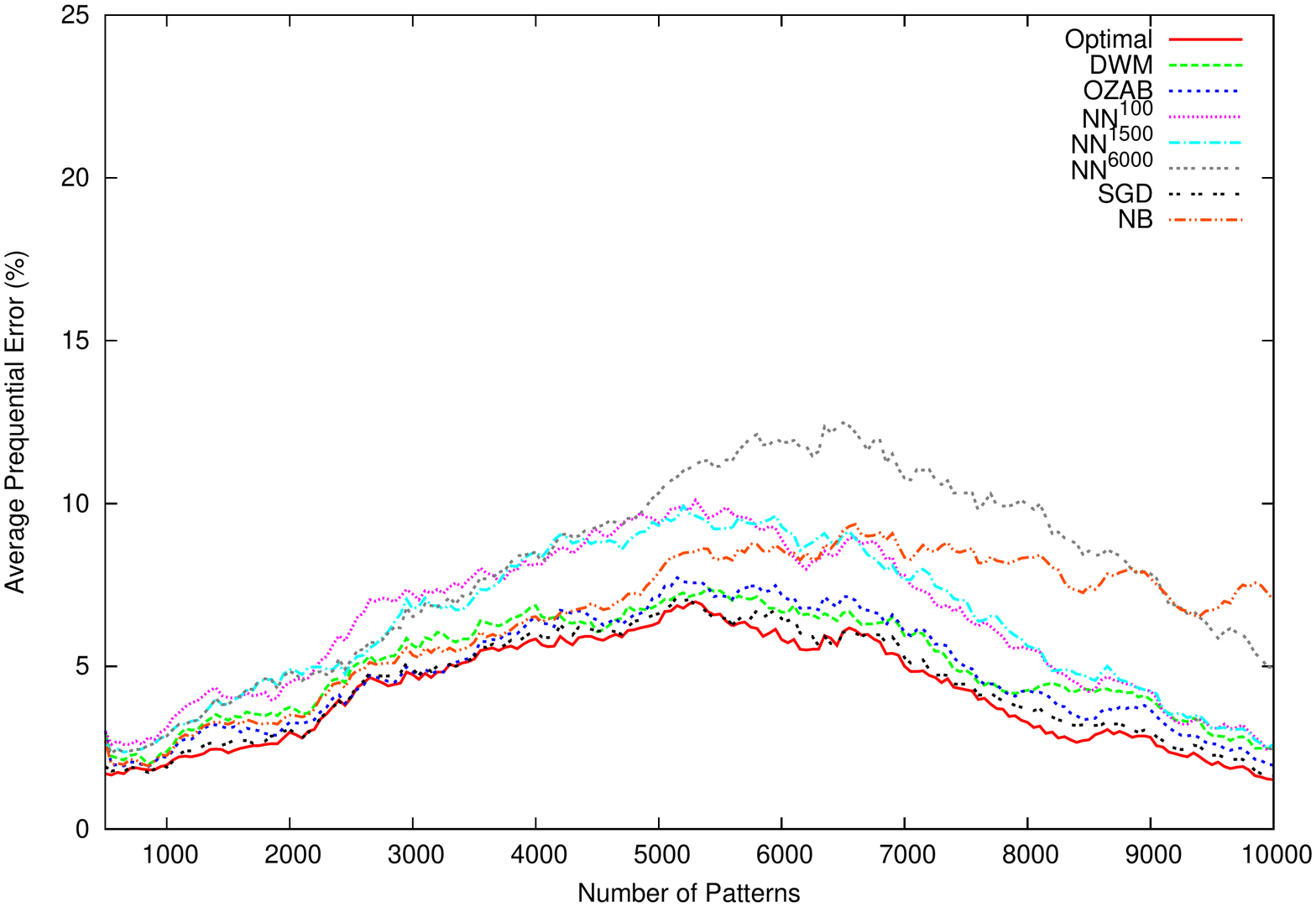}
\label{fig:NSLC:AE500}
}
\\
\subfloat[NSGT-I]{
 \includegraphics[width=0.32\linewidth]{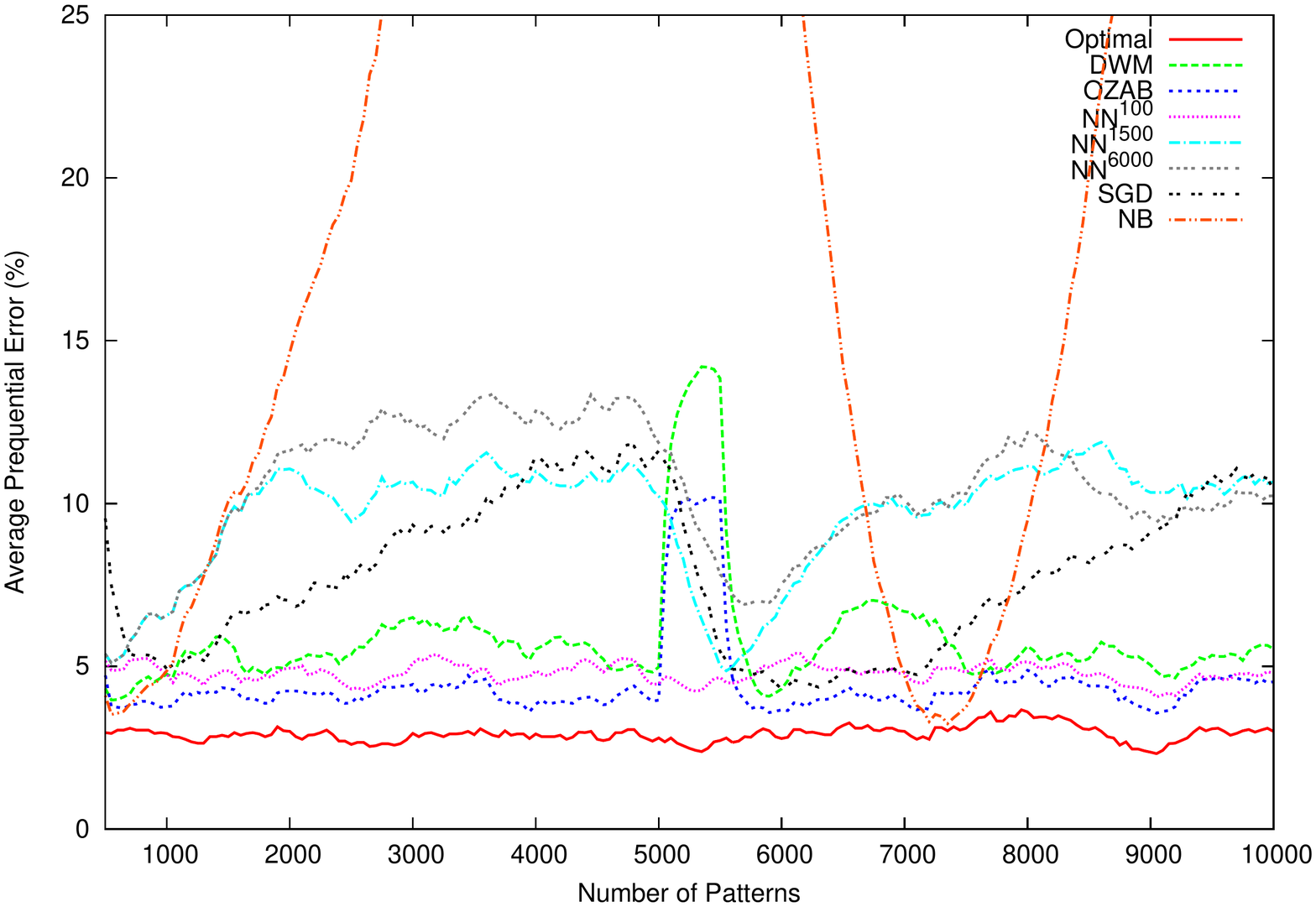}
 \label{fig:NSGT-I:AE500}
 }
\hfil
\subfloat[NSPC]{
\centering
 \includegraphics[width=0.32\linewidth]{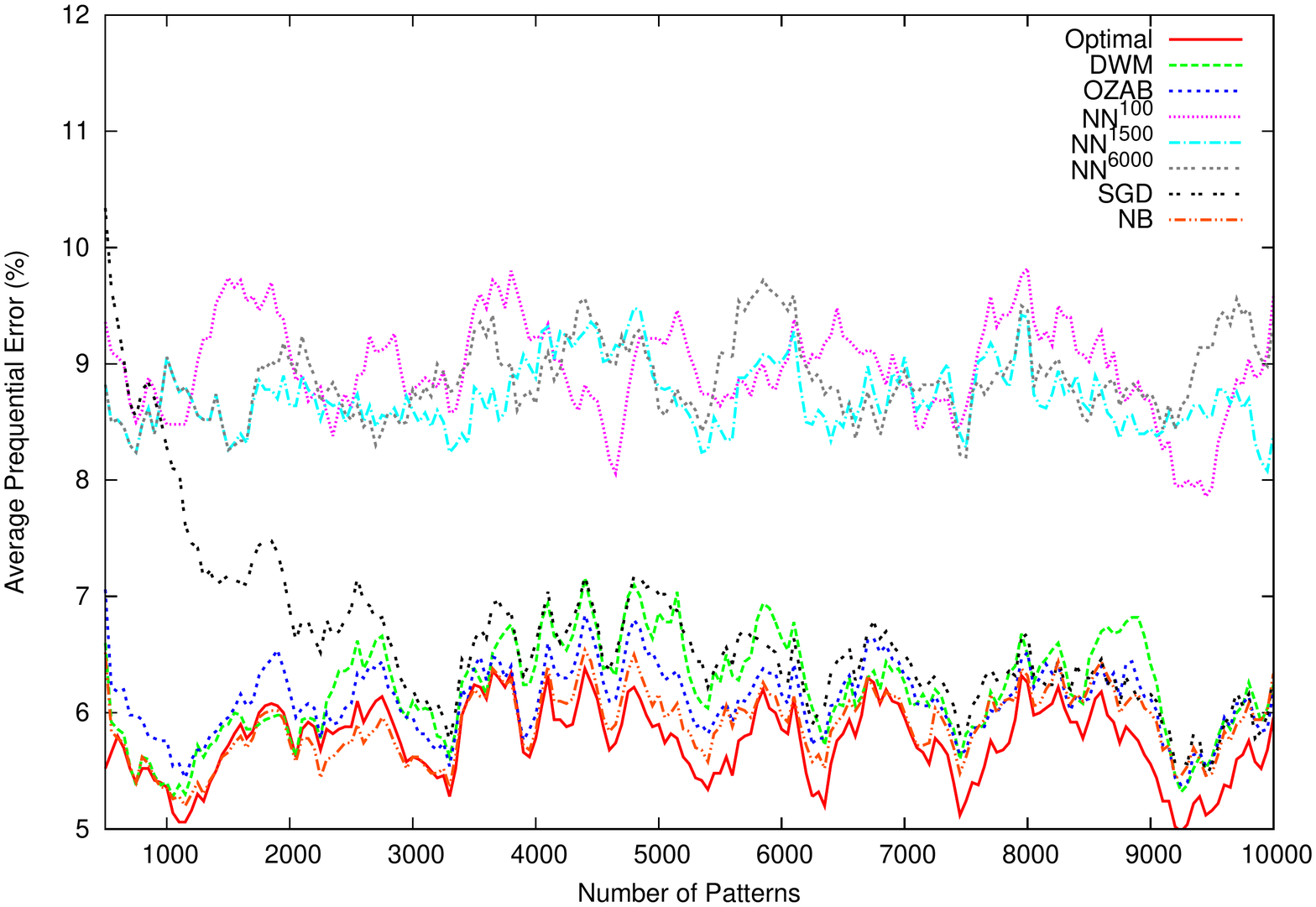}
\label{fig:NSPC-S:AE500}
}
\hfil
\subfloat[NSPC-A]{
\centering
\includegraphics[width=0.32\linewidth]{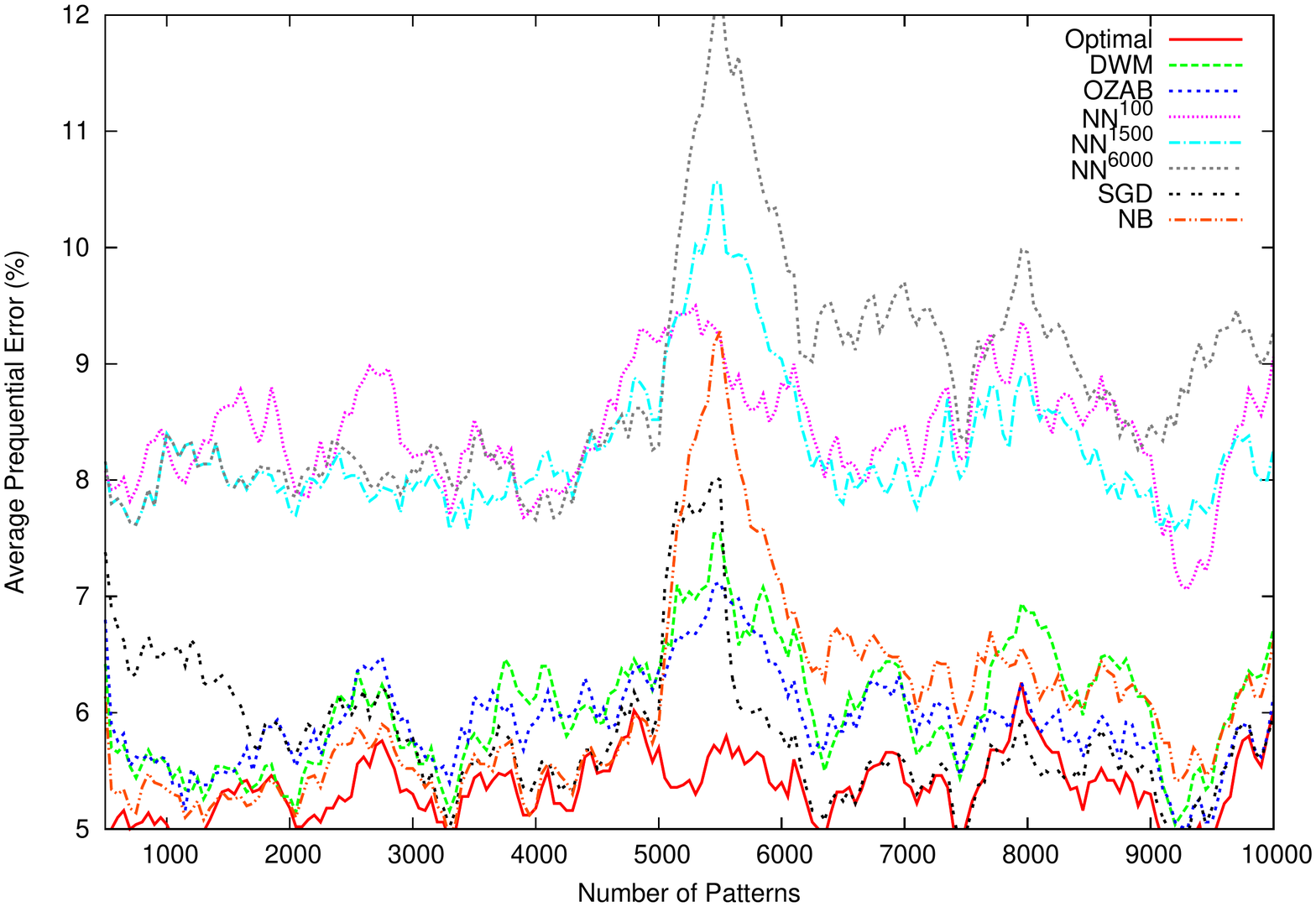}
\label{fig:NSPC-A:AE500}
}
\\
\subfloat[NSGT-F]{
\includegraphics[width=0.32\linewidth]{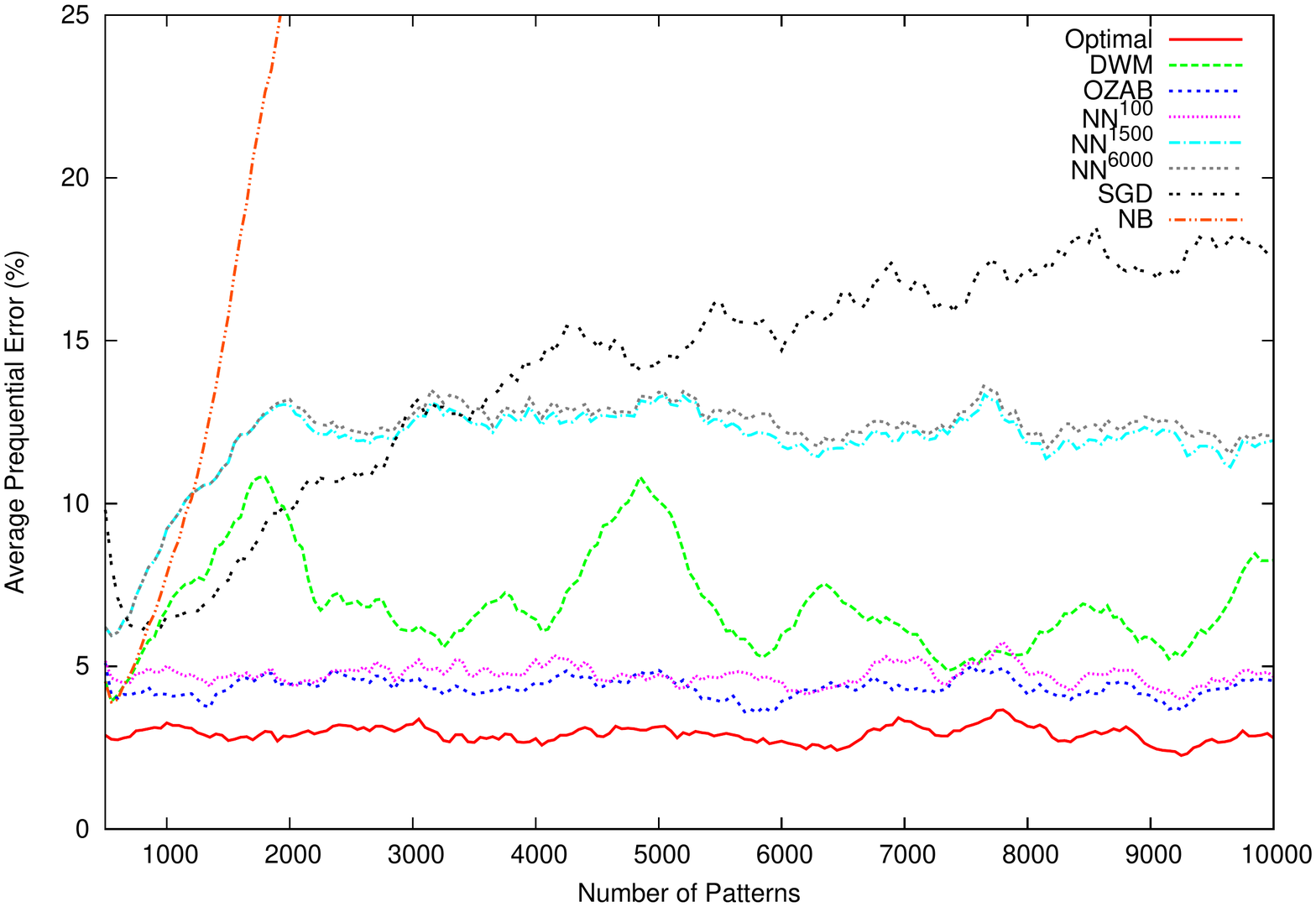}
\label{fig:NSGT-F:AE500}
}
\hfil
\subfloat[NSGT-5D]{
\includegraphics[width=0.32\linewidth]{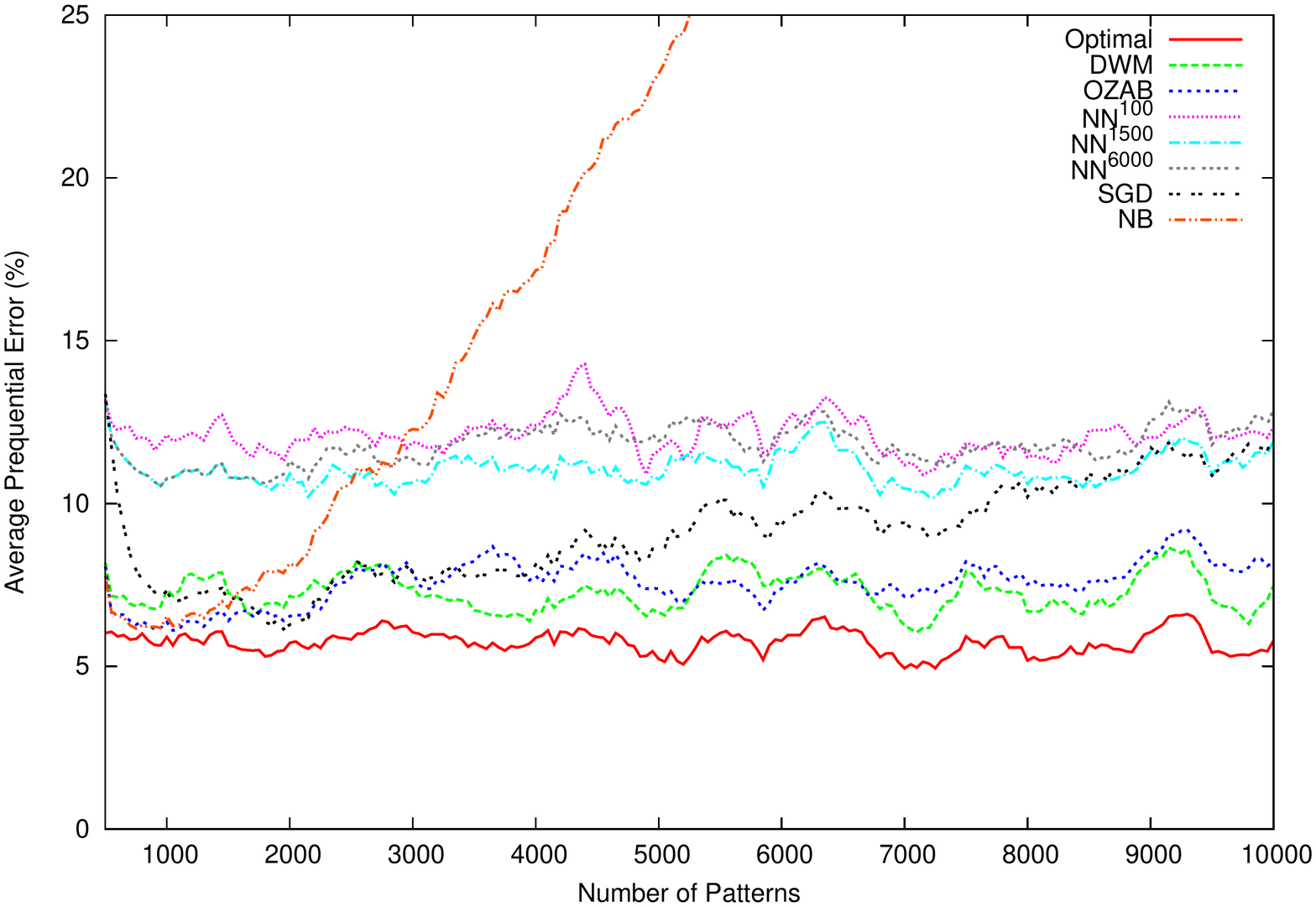}
\label{fig:NSGT-5D:AE500}
}
\hfil
\subfloat[NSCX]{
\includegraphics[width=0.32\linewidth]{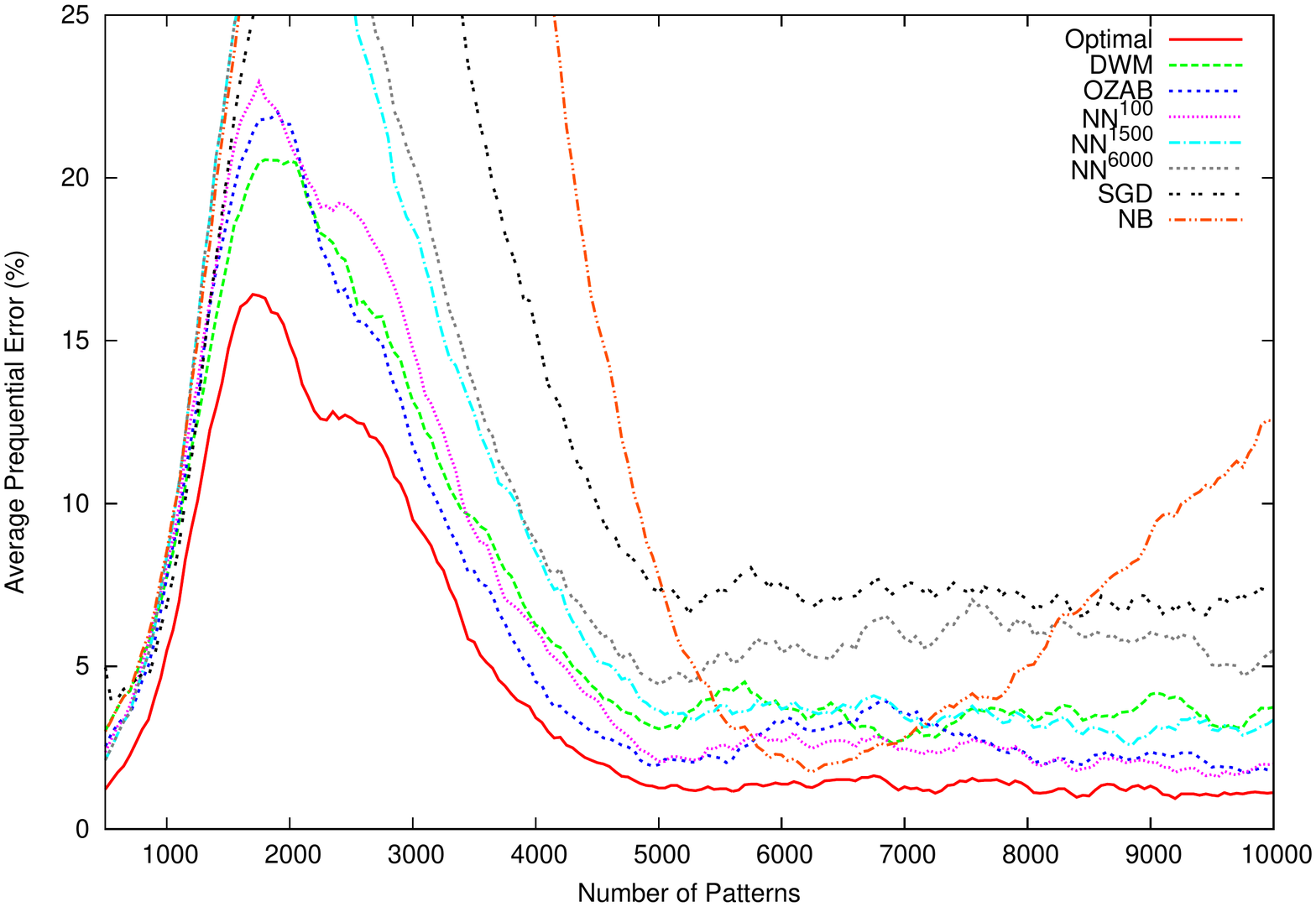}
\label{fig:NSCX:AE500}
}
\caption{Evolution of average prequential error for our datasets, with a
window of 500 patterns ($AE^{500}_{preq}(n)$).}
\label{fig:AllDatasets:AE500}
\end{figure*}

In addition to the final value of $AE_{preq}$, much information may be extracted from
figures that show $AE^{500}_{preq}$. They can be used to show the dynamic
behavior of some of the methods, as they average error during a period of time,
thus providing clear transitory situations. For instance, 
Fig.~\ref{fig:NSGT-S:AE500} shows that NB is not able to adapt to
this type of change at all, as the error increases up to 50\% and never recovers.

% All the plots for all the datasets

With SGD, the error slowly increases over time, so the longer the dataset, the worse for
this method. SGD is creating a separation surface that takes into
account past data that is no longer valid, and in doing so, it is not
really forgetting contradictory data. Ensemble algorithms specifically
designed for non-stationary datasets (DWM and OZAB) perform well. Even though
their approaches to non-stationarity are different, both are working in this
case, and the error tends to be stable after a certain number of patterns.
Also NN$^{100}$ has a stable $AE^{500}_{preq}$, because it avoids contradiction
by taking into account only the most recent data. However, NN is less accurate
than other methods when distributions overlap.

Regarding NSGR, 
Table~\ref{tab:summary}  shows that the optimal Bayes
serious concern to most algorithms as shown by NN$^{100}$; it can be
classified quite accurately by simply adjusting the memory window to a minimal
size, because there is no significant overlap between distributions. In this
dataset SGD achieves a second-best performance, because it seems to be able to
rotate the plane separating both distributions as needed. On the other hand,
having a long-term memory is clearly detrimental, as we can see from NN with a long
window (6000) because if all data is retained in the memory, the new data completely
contradicts past data. NB is the extreme case of this effect.

Some additional information can be obtained from Fig.~\ref{fig:NSGR:AE500}.
When distributions move close to 90 degrees (pattern 2500) new data
contradicts the old data learned at the start. The contradiction reaches a maximum when
angle is 180 degrees (pattern 5000). Both
periods are clearly shown in Fig.~\ref{fig:NSGR:AE500} by the behavior of NB, that is unable to
adapt to this type of contradiction.

Different algorithms react differently in this
moment of change, and the amount of deviation from the original error and the
time it takes to revert to a good level of error may be used as a measure of the
``inertia'' of the algorithms, that is, resistance to change. In checking the
values associated with the peaks starting at pattern index 2500 we can deduce that
OZAB detects change earlier than DWM and adapts slower. DWM detects change
near pattern of index 5000 and is abruptly affected. Then, it rapidly adapts
towards a good classification error. SGD is unaffected by this change, so we
understand that it is adapting constantly to the new data.

\subsection{Contradiction in the Presence of Local Change}
\label{sec:NSLC}

The dataset in this section includes different trajectories for each of the
distributions. We consider these changes as a local transformation in the input
space. In Dataset NSLC, for non-stationary local change, the starting and final
distributions can be easily identified, as they begin separately -- see for
instance Fig.~\ref{fig:NSLC:Plot}.

When the middle of the sequence has been attained, the distributions overlap and then separate
again. Even in the presence of change, the patterns are sampled from distributions that
are located roughly on each side of a vertically-oriented decision frontier.
This means that a complex representation may not provide much advantage over a
very simple one, such as a single linear hyperplane or a NN classifier with a
memory composed of just a few patterns. The optimal frontier, however, should
perform a rotation of 90 degrees during the sequence to properly classify data.

Table~\ref{tab:summary} shows that many algorithms are able to perform
well in the NSLC dataset.The less accurate is again NN with the longest sliding window. The reason
for the performance of NN$^{6000}$ is that a long sliding window size (6000) is unable to forget past knowledge and assigns
the wrong class to new data over the end of the dataset. 

In this case we can assess that sometimes incremental algorithms not designed
for non-stationary data may seem to work well if they are only evaluated using $AE_{preq}$ .
Indeed, NB leads to a result similar to that observed for NN in this dataset because it has a good behavior
until distributions exchange their positions. The best algorithm is SGD, since this
dataset has a simple representation with a surface of separation that 
changes slowly enough for SGD to adjust to contradiction.

A more detailed analysis (Fig.~\ref{fig:NSLC:AE500}) reveals some information that
is not obvious from the accumulated error measure.

Indeed, around pattern 5000 where the overlap is at a maximum, the error rises above 10\% for
NN, because NN is the most affected by the increase in the distribution overlap (due
to the nearest neighbor sensitivity to noise), but remains close to the optimum
for the rest. When the two distributions are separating, all of SGD, DWM and
OZAB track the new distributions following the optimal error very
closely. Also NN$^{100}$ decreases; but the NB error is increasingly separating from
the optimal, indicating clearly that even when the overall accuracy is satisfactory, 
NB is only performing well during the first part of the dataset.

%\newpage

\subsection{Long-term Memory}

We define long-term memory as the ability of an algorithm to retain what was
learned in the past as long as it is not contradictory with more recent learning.
To assess this property, we propose the dataset NSGT-I (non-stationary
global translation, iterated). This dataset has two phases, each of
them based on dataset NSGT. For the first phase, we sampled data for half of
the duration (5000 patterns). Then, distributions were reset to their starting
positions and we repeated the process. 

In Fig.~\ref{fig:NSGT:Plot} we see how 
patterns for the first phase are distributed in two diagonal bands when the
second phase begins. An algorithm with long term memory might be able to learn
the surface of separation and thus improve its results for the second phase of the dataset.

Results in %Table~\ref{tab:NSGT-I:Contrasts} 
Table~\ref{tab:summary} show that OZAB and  
NN$^{100}$ are the best algorithms, and also DWM performs well. However, this
seems contradictory to the fact that NN$^{100}$ has, by definition, no long term
memory. 

The difficulty of deciding if the algorithm exhibits long-term memory is that
the error metric is not an adequate means of evaluating on this feature. Both phases
of the dataset must be compared, which can be done by analysing the average prequential error $AE^w_{preq}$ plotted in
Fig.~\ref{fig:NSGT-I:AE500}.

In this figure we detect different behaviors:
\begin{itemize}
  \item OZAB and DWM have a peak around pattern 5000, and after some learning
  they return to acceptable levels of error. That is, the model is not
  retaining data from the first phase when the second phase begins,
  and some time is required to adapt to the phase transition.
  \item NN$^{100}$ is unaffected by the transition at pattern 5000. This is a
  result of the fact that the algorithm takes into account  the new
  patterns for classification immediately, while old patterns in memory are
  distant from the area where patterns are being observed. That is, the results are not caused by
  long-term memory, but by instant adaptation.
  \item In contrast, SGD seems to retain useful information concerning the first
  part of the algorithm, which accounts for a reduction in error during phase 2. For SGD,
  the error is abruptly reduced when the second phase begins at its minimum level.
  We conclude that SGD exhibits a behavior compatible with long
  term memory.
\end{itemize}

For direct comparison of the effect of long term memory in NN algorithms,
in Fig.~\ref{fig:NSGT-I:AE500} we can compare
results for NN with different window sizes, compared to the error of NB. The plot indicates that NN$^{6000}$
behaved better in the second phase than in the first phase. However, this was
not shown in the average error measure as the overall performance was
clearly better when the window size was smaller (NN$^{100}$).

Also NB exhibits long term memory, as phase 2 has less error than phase 1. 
The error drops until the Bayes error has been reached at the middle of phase 2
(pattern 7000), where distributions have moved half of their trajectory. 
This means that in phase 1 an ``average'' model was learned which was able to perform
optimally at this point in time.

\subsection{Change in Distribution Frequency}
\label{sec:NSPC-S}

The NSPC dataset contains data from three distributions, two of class A (A1
and A2), one of class B. The class B distribution does not change during the dataset. Distributions of
class A undergo a change in a priori probability: at the beginning of the dataset, the
one to the right (A2) has a probability of $0.45$ and the one to the left (A1) has a
probability of $0.05$
(Fig.~\ref{fig:NSPC-S:Plot:Start}). 
During the
dataset, probabilities are slowly exchanged during a time period of 9000 patterns, until the situation is reversed towards the end of the dataset
(Fig.~\ref{fig:NSPC-S:Plot:End}). 
As a priori
probabilities for both distributions for class A always sum up to $0.5$, both
classes (A and B) are balanced at all times during the dataset.

Table~\ref{tab:summary} shows results for this dataset. In
Fig.~\ref{fig:NSPC-S:AE500} we present the average prequential error ($AE^{500}_{preq}$) for a sliding
window of 500 patterns.

In this dataset NB has the best performance, meaning that the slow variation in
a priori probabilities is processed adequately and incorporated into its
model. Nearest-neighbor approaches lead to the
worst results, independently of the window size; however a medium window size
(NN$^{1500}$) seems to provide a slightly better result in these conditions.
In addition to NB, the algorithms DWM, OZAB and SGD also follow the optimal Bayes classifier
closely.
Note that the performance of DWM is much poorer than that of NB, even when its base classifier is NB.
This means that the ensemble mechanism is being confused by this dataset.

\subsection{Appearance of New Concepts}
\label{sec:NSPC-A}

The NSPC-A dataset is a variation on NSPC. A priori
probabilities are not changed slowly, but abruptly, from a starting
situation where priors are $0$ for A1 (the leftmost distribution) and
$0.5$ for A2 (the rightmost distribution), to the end situation with the
opposite situation of $0.5$ (for A1) and $0.0$ (for A2) -- see illustration in 
%Fig.~\ref{fig:NSPC-A:Plot}.
Fig.~\ref{fig:NSPC-S:Plot:Start}, Fig.\ref{fig:NSPC-S:Plot:5000} and
Fig.~\ref{fig:NSPC-S:Plot:End}.

Table~\ref{tab:summary} shows results for this dataset, while
Fig.~\ref{fig:NSPC-A:AE500} presents the  average prequential error for
a sliding window of 500 patterns.

The behavior of algorithms for this dataset depends on how they react to the abrupt
change, more than the actual difference in the accumulated error, which is
similar for all algorithms, except for those based on NN. SGD and OZAB are
significantly better than NB, but the difference in accumulated error is small anyway.

Fig.~\ref{fig:NSPC-A:AE500} indicates that most of the algorithms perform well during the
two ``stable'' periods at the beginning and the end of the dataset, while
reaction to the change in pattern 5000 is very different: NB has a significant
increase in error and adapts very slowly towards the end of the dataset; SGD has
a high increase, but adapts quickly; DWM has an even smaller increase at the
peak, but seems to be unable to adapt completely, since the performance at the end of the
dataset is poorer than at the beginning; OZAB exhibits the smallest
increase in error, but returns to the levels it had at the beginning.

As with NSPC, nearest-neighbor approaches have the worst results. In 
Fig.~\ref{fig:NSPC-A:AE500} we see that for NN algorithms, the
lack of performance is not only related to the transition in pattern 5000, but
to problems in separating these distributions that have a high degree of overlap.

\subsection{Speed of Change}

In this section we investigate the effect of the speed of change in terms of accuracy. For
that purpose, we use the dataset NSGT-F (a ``fast'' version of NSGT), and 
compare the results with results of the basic (or ``slow''
version) shown previously. NSGT-F is described by two distributions that undergo 
a global translation (see Fig.~\ref{fig:NSGT:Plot}) as in NSGT, but distribution
centers are moved three times the distance in each coordinate. The speed of change is thus
three times the speed in NSGT. 

In Table~\ref{tab:summary}, comparison between NSGT-F and the NSGT dataset shows
that NB, SGD and DWM have a very significant decrease in performance, while OZAB is mostly unaffected by this change. NN$^{100}$ performs better in this case, almost
matching the performance of OZAB.

Fig.~\ref{fig:NSGT-F:AE500} provides an illustration of
the average prequential error for a sliding window of 500
patterns. The trends for NB and SGD were the same as for NSGT, i.e., their
error increased towards the end of the dataset. Higher velocity has an
important effect because this increase starts earlier and grows at a faster rate.
The decrease in accuracy for DWM (which performed well when translation was
slow), is caused by peaks of increased error that are later corrected. This can be explained
by the mechanism of member creation of the DWM ensemble. In this case, 
too much error is accumulated until a new ensemble member is activated
to adapt to the new data.

\subsection{Effect of Dimensionality}

Datasets in two dimensions allow visualization, and are thus convenient for
inspection and validation of performances. However, behavior of algorithms may
change drastically as dimensionality increases given the exponential growth of the
input space this generates.

In some research studies, the effect of a dimensionality increase is verified by adding noisy
and useless dimensions that must be filtered out by learning methods in order to
achieve optimal classification accuracy. We are not considering this case here
as we believe this to be a problem of feature selection/extraction, which
may be dealt with using other techniques designed for that purpose. Instead, we
are including additional dimensions that are useful and informative, with
changes over the distributions of the data that are performed over all these
dimensions.

The NSGT-5D dataset is based on the NSGT dataset presented in
Sec.~\ref{sec:ContradictionGlobalChange}, with distributions this time defined
in a 5D input space and a translation of the distributions over the time steps
parallel to the vector $[1.0\,1.0\,1.0\,1.0\,1.0]$.  

The 2-D projection of the distributions and their movement
would be similar to that of NSGT (see Fig.~\ref{fig:NSGT:Plot}).
 
The covariance matrices parametrizing the distributions have been adjusted to
ensure that there is a significant overlap between the distributions. The amount of
overlap was estimated by the error of the optimal Bayes classifier and tuned to 
be similar to the one reported for NSGT. We used symmetrical distributions for
simplicity.
 
It is expected that some algorithms require more patterns to be able to model a
distribution as dimension grows. However, increasing the number of patterns
would decrease the speed at which the distributions are moving in the input
space. We adjusted the distance moved by the distribution centers to ensure that the dataset
NSGT-5D use the same number of patterns ($10000$), and that the speed of
movement remains equal to the speed in NSGT.

Table~\ref{tab:summary}
presents the results obtained for NSGT-5D, while
Fig.~\ref{fig:NSGT-5D:AE500} presents the evolution of its prequential error over time.

These results are similar to the results obtained for NSGT, except for the fact that
the performance of the NN algorithms decreased for NN$^{100}$, while
NN$^{6000}$ remained unaffected. This is expected due to the known dimensionality
issues of the NN rule. This means that in these types of algorithms the window
size must be adjusted to take into account both change and dimensionality at
the same time. 

As in the case of NSGT, the algorithms DWM and OZAB exhibited better performance than that of SGD, especially if we consider the last part of the dataset where SGD is increasing its error. So
even with an increase in dimensions, the former results are generally applicable to
these algorithms.

\subsection{Complex Non-stationary Changes}

Finally, we introduce the dataset NSCX that combines several types of changes.
This dataset has two initial distributions
(A1 and A2) (Fig.~\ref{fig:NSCX:Plot:Start}) that are
then separated by a third one (B) which appears gradually (change in prior) and
translates downwards (local change) while A2 moves diagonally. A1 and A2
exchange their a priori probabilities gradually. We also perform a rotation of 
distributions A1 and B, and the variance of A1 grows with time. Note that, contrary to the previous datasets,
during part of the evolution of the dataset, distributions are non-separable by 
a single hyperplane (Fig.~\ref{fig:NSCX:Plot:5000}), but
towards the end they can easy be separated with a linear frontier
(Fig.~\ref{fig:NSCX:Plot:End}).

Table~\ref{tab:summary} presents the results obtained for the NSCX dataset,
while Fig.~\ref{fig:NSCX:AE500} presents the evolution of its prequential error over time.

Overall, OZAB is the best algorithm followed by DWM and NN$^{100}$ which are able to
provide good results. However, SGD is unable to perform much better than
NB. It must be noted that OZAB and NN are the algorithms that, in general
terms, are able to classify distributions with more complex representations.

The plot for the optimal error (see Fig.~\ref{fig:NSCX:AE500}) indicates that the
distributions increasingly overlap during the first part of the dataset but then become
separated again. This improves the overall behavior of the NN
algorithms towards the end. While the final situation seems to be an ``easy''
situation, that is, it is linearly separable
(Fig.\ref{fig:NSCX:Plot:End}), SGD is not able 
to evolve towards a linear separation frontier. At the end of the dataset, DWM
shows no adaptation towards the optimal classifer error.

\section{Summary of Results}

For a proper assessment of Non-Stationary Learning (NSL) methods, we are proposing a methodology that includes:
several datasets that test different features of the methods;
datasets with parameters that allow the speed of change to be varied;
metrics that are able to discard averaging effects of the accumulated error rate (i.e., average prequential error with a sliding window).

The NSGT dataset indicates that, as expected, NB is not designed
to adapt to even global change, that is, the simplest type of
change involving global displacement in the input space with no relative
change. For SGD we showed that error increases with time. OZAB
and DWM can deal with contradiction by adapting their internal model, either
instance weights (OZAB) or member weights (DWM) when their accuracy is reduced. 
Also NN$^{100}$ achieves good results with a simple time-based forgetting
mechanism.

NSGR shows that if a very simple representation is enough to represent
data (SGD, NN$^{100}$), then simple algorithms may work better than complex
ones (OZAB, DWM). 

For local change (NSLC), SGD was shown as a good option that was able
to adapt to transitory increase in complexity. It can also generate a very
exact linear separation between overlapping distributions that can rotate as
the distributions intersect. NN approaches, on the other side, are much less
tolerant to overlap and are affected in a greater proportion when the Bayes error increases.

The NSGT-I dataset was  specifically designed to test which methods
can recall models created early during learning. We were able to detect this
feature in NB, SGD, and in the version of NN that had a longer
memory window. 

Change in a priori probability was well processed by all algorithms when it was
gradual (NSPC). However, abruptly introducing change
(NSPC-A) induces different behaviors depending on the inertia and adaptability 
of the different algorithms. Here, OZAB was very successful but slow to adapt,
while SGD was greatly perturbed by the change, but was able to rapidly adapt later.

Concerning speed of change, the behavior of the analyzed algorithms
can be evaluated using our methodology. For instance, when changes occurs more rapidly
(NSGT-F dataset), it appears that SGD is unable to
retain its good results due to its accumulative increase in error.
Also DWM seems to have more difficulty in adapting to rapid change than OZAB, while for NN algorithms
this depends on the window size: as expected, a small window size is adequate for
rapid adaptation.

The dataset NSGT-5D, with its extra dimension, also shows some of the well-known
features of the algorithms. Behavior in two dimensions does not always carry-on
to input spaces of higher dimensions, specially for NN, where the smaller window
size, that was capable of adaptating very well to change in many situations, is
unable to maintain the good results it has for the 2-dimensional version of
this dataset.

Finally, the dataset NSCX can be used to
evaluate representation issues and complex situations. In this context NN
seemed to be a good option, except when overlap between distributions of two
classes increases. The SGD algorithm relies on a linear discriminant, and thus would
not be able to cope with non-linearly separable datasets such as NSCX. 
This weakness is clear in Fig.~\ref{fig:NSCX:AE500} where SGD was
unable to perform well once the third distribution was introduced. Both OZAB and
DWM were able to perform reasonably well.

Overall OZAB and DWM were the most flexible and accurate NSL algorithms.
They have an equilibrium between error (usually lower for OZAB) and speed
(DWM has a much lower computational cost). SGD was able to perform well when
the representation of the separation surface is simple and change is slow. NN
with a short memory window can be competitive and is flexible enough for any
type of surface of separation or velocity of change, but it is adversely
affected both by noise (overlap between distributions) and the dimensionality of the problem.

\section{Conclusions}

This paper presents a testbed for the evaluation of algorithms in the field of Incremental Learning from Non-Stationary data (NSL). This task is defined in relation to other fields that share some characteristics. Specifically we refer to these algorithms as incremental learning algorithms and they must deal with sequential data, continuous flow, and non-stationary distributions. 

This field has no definite criteria for comparison. The analysis of the state of the art shows that authors are comparing algorithms using datasets that were not initially oriented towards NSL. Also, the basic characteristics of these datasets such as the amount and speed of change remain undefined. This means that we cannot determine which features of an algorithm are being tested when we claim that a method provides good results on one of the datasets in the literature.

Most of our work has been oriented towards the identification of the most relevant tests that have to be performed in order to determine if a proposed incremental learning algorithm will be able to perform the task at hand. These tests include contradiction, long-term memory, local distribution invariance, global distribution invariance, data frequency invariance, adaptation to new data, dimensionality invariance, and velocity invariance.

We have created artificial datasets where each of the former tests can be performed independently. Experiments have been performed using these datasets and well-known algorithms. Results confirm that our testbed can be used to compare candidate algorithms in a generic way. The discussion of the results also provides guidelines on how metrics should be interpreted.

\ifCLASSOPTIONcompsoc
  \section*{Acknowledgments}
\else
  \section*{Acknowledgment}
\fi

This article has been funded by the Spanish Ministry of Science and Innovation (project MOVES, TIN2011-28336) and NSERC-Canada. We thank Annette Schwerdtfeger for proofreading this manuscript.

%\bibliographystyle{IEEEtran}
%\bibliography{IEEEabrv,gpsc}

\begin{IEEEbiography}[{\includegraphics[width=1in,height=1.25in,clip,keepaspectratio]{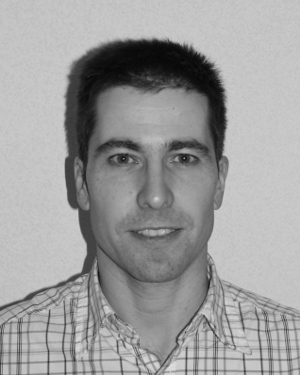}}]{Alejandro
Cervantes} graduated as Telecommunications Engineer at
 Universidad Polit\'{e}cnica of Madrid (Spain), in 1993. He received 
 his PhD in Computer Science at Carlos III of Madrid in 2007.  He is
 currently an assistant professor at the Computer Science
 Department at this same University. His current
 interests focus in algorithms for classification of non-stationary data, large
 multi-objective optimization problems, and swarm intelligence algorithms for
 data mining.
 \end{IEEEbiography}
 
\begin{IEEEbiography}[{\includegraphics[width=1in,height=1.25in,clip,keepaspectratio]{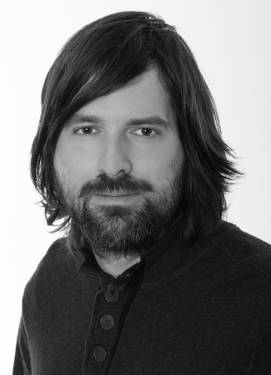}}]{Christian
Gagn\'{e}} received a B.Ing. in Computer Engineering and a PhD in Electrical Engineering
from Université Laval in 2000 and 2005, respectively. He is professor of
Computer Engineering at Université Laval since 2008. His research interests are
on the engineering of intelligent systems, in particular systems involving
machine learning and evolutionary computation. He is member of the editorial
board of Genetic Programming and Evolvable Machines, and participated in
the organization of several conferences.
\end{IEEEbiography}

\begin{IEEEbiography}[{\includegraphics[width=1in,height=1.25in,clip,keepaspectratio]{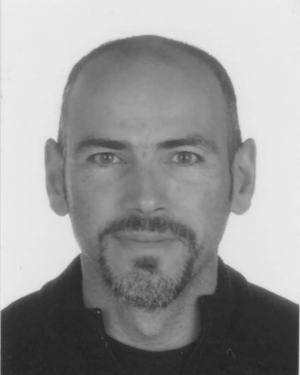}}]{Pedro
Isasi} 
is engineer in Computer Science for Polytechnic University of Madrid
since 1991 and received his PhD at this same university in 1994. From 2001, he is a
Professor in Computer Science at Carlos III of Madrid University. He is funder
and director of the Neural Network and Evolutionary Computation Laboratory. His
principal researches are in the field of Machine Learning, Metaheuristics and
evolutionary optimization methods, mainly applied to prediction and
classification.
\end{IEEEbiography}

\begin{IEEEbiography}[{\includegraphics[width=1in,height=1.25in,clip,keepaspectratio]{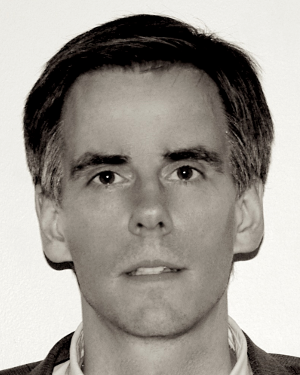}}]{Marc
Parizeau} Marc Parizeau is a professor of Computer Engineering at Université Laval,
Québec City. He obtained his Ph.D. in 1992 from École Polytechnique de
Montréal. His research interests are %mainly
in the field of intelligent
systems, in machine learning for pattern recognition in particular, %as well as in 
parallel and distributed systems. In 2008, he created a High Performance
Computing %(HPC) 
center at Université Laval, and is the current scientific
Director of Calcul Québec, %an HPC consortium for the province of Québec, also
one of the four regional divisions of Compute Canada, the national HPC
platform.
\end{IEEEbiography}

\end{document}